\documentclass[acmsmall]{acmart}

\AtBeginDocument{%
  \providecommand\BibTeX{{%
    \normalfont B\kern-0.5em{\scshape i\kern-0.25em b}\kern-0.8em\TeX}}}

\setcopyright{acmcopyright}

\usepackage{bm}
\usepackage{multirow}
\usepackage{utfsym}
\usepackage{tabularx}
\usepackage{array}
\usepackage{longtable}
\begin{document}

\title{Multi-Agent Reinforcement Learning: Methods, Applications, Visionary Prospects, and Challenges}

\author{Ziyuan Zhou}
\author{Guanjun Liu} 
\authornote{corresponding author}
\affiliation{
  \streetaddress{Department of Computer Science}
  \institution{Tongji University}
  \city{Shanghai}
  \country{China}}
\email{{ziyuanzhou ,liuguanjun}@tongji.edu.cn}

\author{Ying Tang}
	\affiliation{
		\streetaddress{Department of Electrical and Computer Engineering}
		\institution{Rowan University}
		\city{Glassboro}
		\state{New Jersey}
		\country{USA}
		\postcode{08028}
	}
	\email{tang@rowan.edu}

\renewcommand{\shortauthors}{Z. Zhou, G. Liu and Y. Tang}
\begin{abstract}
Multi-agent reinforcement learning (MARL) is a widely used Artificial Intelligence (AI) technique. However, current studies and applications need to address its scalability, non-stationarity, and trustworthiness. This paper aims to review methods and applications and point out research trends and visionary prospects for the next decade. 
First, this paper summarizes the basic methods and application scenarios of MARL. 
Second, this paper outlines the corresponding research methods and their limitations on safety, robustness, generalization, and ethical constraints that need to be addressed in the practical applications of MARL.
In particular, we believe that trustworthy MARL will become a hot research topic in the next decade. 
In addition, we suggest that considering human interaction is essential for the practical application of MARL in various societies.
Therefore, this paper also analyzes the challenges while MARL is applied to human-machine interaction. 
\end{abstract}




\maketitle

\section{Introduction}
\begin{table}
\centering
\caption{The difference between this paper and other related reviews.}
\label{Tdiff}
\begin{tabular}{|c|cccccc|}
\hline
\multirow{2}{*}{Work}    & \multicolumn{5}{c|}{Scope}   \\ \cline{2-6} 
                         & \multicolumn{1}{c|}{SARL} & \multicolumn{1}{c|}{MARL} & \multicolumn{1}{c|}{Applications} & \multicolumn{1}{c|}{Trustworthy}  & \multicolumn{1}{c|}{Human} \\ \hline
\cite{kapoor2018multi, Wong2022, hernandez2018multiagent} & \multicolumn{1}{c|}{\usym{2713}}  & \multicolumn{1}{c|}{\usym{2713}}    & \multicolumn{1}{c|}{\usym{2613}}            & \multicolumn{1}{c|}{\usym{2613}}    &\multicolumn{1}{c|}{ \usym{2613} }   \\ \hline

\cite{9738819} & \multicolumn{1}{c|}{\usym{2713}}  & \multicolumn{1}{c|}{\usym{2713}}    & \multicolumn{1}{c|}{Future Internet}            & \multicolumn{1}{c|}{\usym{2613}}       &\multicolumn{1}{c|}{ \usym{2613} }   \\ \hline

\cite{9043893} & \multicolumn{1}{c|}{\usym{2713}}  & \multicolumn{1}{c|}{\usym{2713}}    & \multicolumn{1}{c|}{\usym{2713}}            & \multicolumn{1}{c|}{\usym{2613}}      &\multicolumn{1}{c|}{ \usym{2613} }   \\ \hline

\cite{wang2022model} & \multicolumn{1}{c|}{\usym{2613}}  & \multicolumn{1}{c|}{Model-based}    & \multicolumn{1}{c|}{\usym{2613}}            & \multicolumn{1}{c|}{\usym{2613}}     & \multicolumn{1}{c|}{\usym{2613}}    \\ \hline

\cite{cui2022survey} & \multicolumn{1}{c|}{\usym{2713}}  & \multicolumn{1}{c|}{Large Population}    & \multicolumn{1}{c|}{\usym{2713}}            & \multicolumn{1}{c|}{\usym{2613}}      & \multicolumn{1}{c|}{\usym{2613}}   \\ \hline

\cite{Zhang2021} & \multicolumn{1}{c|}{\usym{2713}}  & \multicolumn{1}{c|}{Decentralized}    & \multicolumn{1}{c|}{\usym{2613}}            & \multicolumn{1}{c|}{\usym{2613}}      & \multicolumn{1}{c|}{\usym{2613}}   \\ \hline

\cite{zhu2022survey} & \multicolumn{1}{c|}{\usym{2713}}  & \multicolumn{1}{c|}{Communication}    & \multicolumn{1}{c|}{\usym{2613}}            & \multicolumn{1}{c|}{\usym{2613}}      & \multicolumn{1}{c|}{\usym{2613}}   \\ \hline

\cite{grimbly2021causal} & \multicolumn{1}{c|}{\usym{2713}}  & \multicolumn{1}{c|}{Causal}    & \multicolumn{1}{c|}{\usym{2613}}            & \multicolumn{1}{c|}{\usym{2613}}      & \multicolumn{1}{c|}{\usym{2613}}   \\ \hline

\cite{gu2022review} & \multicolumn{1}{c|}{\usym{2713}}  & \multicolumn{1}{c|}{\usym{2713}}    & \multicolumn{1}{c|}{\usym{2613}}            & \multicolumn{1}{c|}{Safety}      & \multicolumn{1}{c|}{\usym{2713}}   \\ \hline

\cite{9536399} & \multicolumn{1}{c|}{\usym{2713}}  & \multicolumn{1}{c|}{\usym{2713}}    & \multicolumn{1}{c|}{\usym{2613}}            & \multicolumn{1}{c|}{Robustness}      & \multicolumn{1}{c|}{\usym{2613}}   \\ \hline

\cite{9308468, electronics9091363, 10.1613/jair.1.14174} & \multicolumn{1}{c|}{\usym{2713}}  & \multicolumn{1}{c|}{\usym{2713}}    & \multicolumn{1}{c|}{\usym{2613}}            & \multicolumn{1}{c|}{Generalization}      & \multicolumn{1}{c|}{\usym{2613}}   \\ \hline

\cite{trustRL} & \multicolumn{1}{c|}{Comprehensive}  & \multicolumn{1}{c|}{Brief}    & \multicolumn{1}{c|}{\usym{2613}}            & \multicolumn{1}{c|}{\usym{2713}}      & \multicolumn{1}{c|}{\usym{2713}}   \\ \hline

Ours & \multicolumn{1}{c|}{Brief}  & \multicolumn{1}{c|}{Comprehensive}    & \multicolumn{1}{c|}{\usym{2713}}            & \multicolumn{1}{c|}{\usym{2713}}      & \multicolumn{1}{c|}{\usym{2713}}   \\ \hline

\end{tabular}
\end{table}

Reinforcement Learning (RL) is extensively explored due to its tremendous potential in solving sequence decision tasks \cite{dqn,doubleq,duelq,ac, a3c,trpo,ppo,ddpg,REINFORCE}. Kaelbling et al. pointed out in 1996 \cite{1996Reinforcement} that RL will be widely used in game playing and robotics. Mnih et al.  \cite{dqn2013} propose Deep Reinforcement Learning (DRL) to combine reinforcement learning with reasoning ability and Deep Learning (RL) with representative capacity, and the performance of the trained agent outperformed that of human players in various Atari games.
Silver et al. use RL to solve Go games in  2007\cite{2007Go} and propose AlphaGo leveraging deep neural networks and Monte Carlo tree search in 2016 \cite{2016AlphaGo}. In robotics, DRL also achieves outstanding developments such as quadrupedal movement  \cite{quadrupedal1, quadrupedal2}. The latest ChatGPT is well-known worldwide and makes use of RL-related technology. In the 20 years since DRL was proposed, there has been a continuous rise in research interest in games and robotics. Visionary applications of RL are summarized in \cite{1996Reinforcement}.

Multi-Agent Reinforcement Learning (MARL) research is advancing significantly based on the issues of poor scalability and non-stationary and has shown remarkable success in a range of applications. We summarize the relevant research on MARL in nine domains, involved in engineering and science. 

However, despite the impressive achievements, it is still necessary to construct trustworthy MARL to apply it to real-world tasks better. Consequently, one of the most critical topics we need to focus on in the next 10 to 20 years is \emph{how to establish a trustworthy MARL}. As stated in \cite{trustRL}, the intrinsic safety, robustness, and generalization of RL still need to improve, making it challenging to realize accurate general intelligence. 
While it mainly focuses on the single-agent domain. Compared to Single-agent Reinforcement Learning (SARL), MARL requires consideration not only of individual policy trustworthiness but also of the reliability of team interaction policies. As the number of agents increases, the complexity of team policies also increases, which increases the difficulty of researching trustworthy MARL. Currently, there is a portion of research on trustworthy MARL, but it is still in the early stages. To promote the development of this field, we conduct a comprehensive investigation of trustworthy multi-agent reinforcement learning from four aspects, including safety, robustness, generalization, and learning with ethical constraints.

By integrating human aspects, it is necessary to take into consideration not just agent collaboration but also the interaction between intelligent physical information systems and human civilization. In relation to MARL for human-machine interaction, we present four challenges: non-Markovian due to human intervention, diversity of human behavior, complex heterogeneity, and scalability of multi-human and multi-machines.

\begin{figure}
    \centering
    \includegraphics[width=0.95\textwidth]{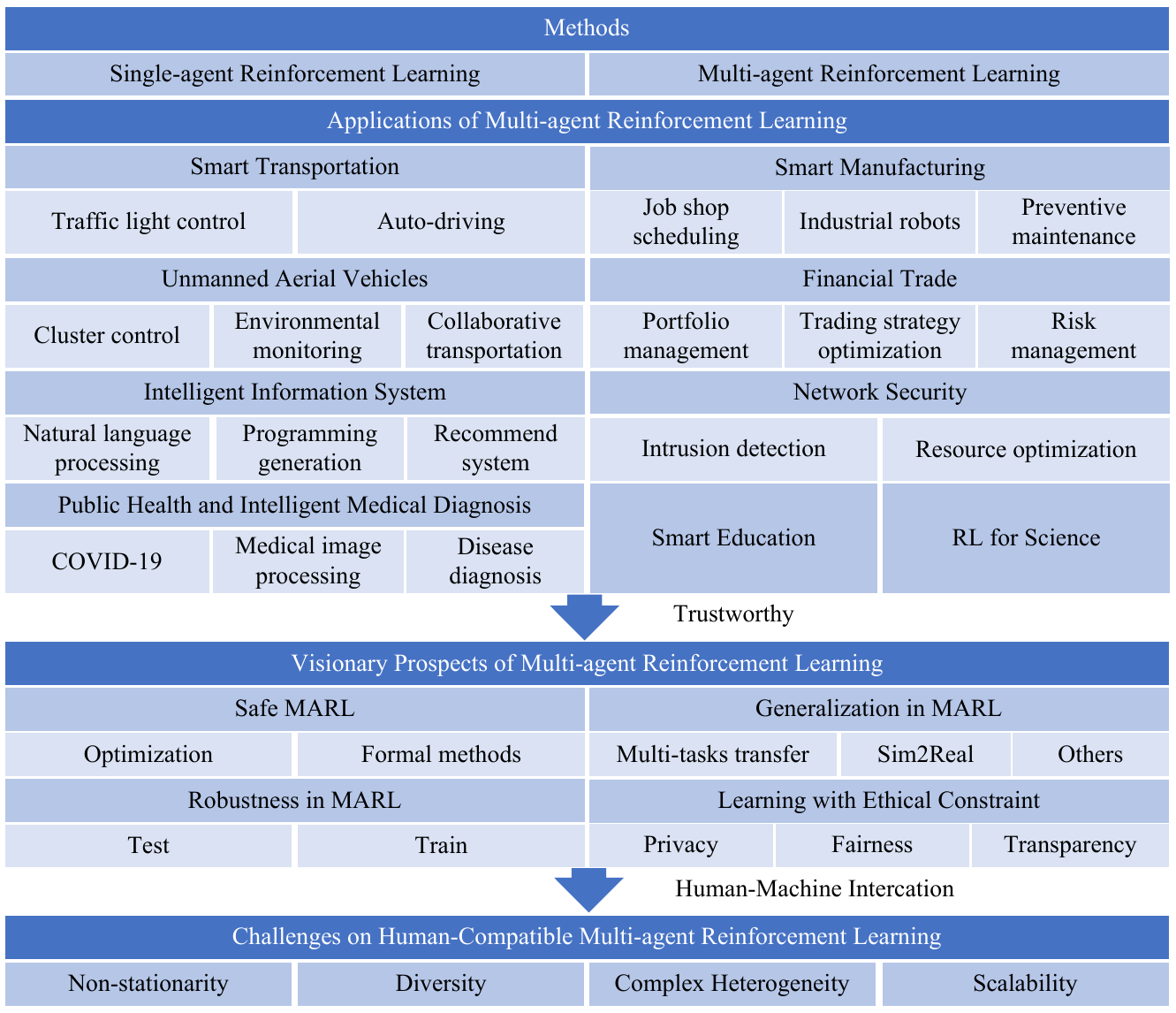}
    \caption{The outline of this survey}
    \label{outline}
\end{figure}

The difference between this paper and other related reviews are listed in Table \ref{Tdiff}. The outline of this paper is shown in Fig. \ref{outline}. The rest of this survey is organized as follows. In Section \ref{Preliminary}, we give a relevant definition of MARL and summarize typical research methods. Section \ref{Applications} shows the specific application scenarios of MARL. Section \ref{Visionary} summarizes the definition, related research, and limitations of trustworthy MARL. In Section \ref{Challenges}, we point out the challenges faced by human-compatible MARL. Section \ref{Conclusion} concludes the whole paper. 
\section{Methods} \label{Preliminary}

\subsection{Single-agent Reinforcement Learning}

The RL agent aims to maximize the total discounted expected reward by trial-and-error interactions with the environment. Markov Decision Process (MDP) helps define models for sequential decisions.
\begin{definition}[MDP]
A Markov decision process can be formulated by a 5-tuple $\left<\mathcal{S}, \mathcal{A}, R, p, \gamma\right>$, where $\mathcal{S}$ is the environmental-state set, $\mathcal{A}$ is the space of agent actions, $R:\mathcal{S}\times\mathcal{A}\times\mathcal{S}\rightarrow\mathbb{R}$ is the reward obtained by the agent for transition to state $s’$ by doing action $a$ in state $s’$, $\mathbb R$ is the set of real numbers, $p: \mathcal{S}\times\mathcal{A}\rightarrow \Delta\left(\mathcal{S}\right)$ is the transition probability from state $s\in \mathcal{S}$ to state $s' \in \mathcal{S}$ given the action $a$, and $\gamma \in \left[0,1\right]$ is the discount factor over time. 
\end{definition}
Solving MDP is to learn a policy $\pi:\mathcal{S}\rightarrow \Delta\left(\mathcal{A}\right)$ that maximizes the expected reward over time, where $\Delta\left(\cdot\right)$ is the probability simplex. The state-action (Q-function) and value functions are as 
\begin{equation} \label{q}
    Q_{\pi}\left(s, a\right) = \mathbb E_{\pi}\left[\sum_{t=0}^\infty \gamma^tR\left(s_t,a_t,s_{t+1}|a_0 = a, s_0 = s\right)\right],
\end{equation}
\begin{equation} \label{v}
    V_{\pi}\left(s\right) = \mathbb E_{\pi}\left[\sum_{t=0}^\infty \gamma^tR\left(s_t,a_t,s_{t+1}|s_0 = s\right)\right],
\end{equation}
where $R\left(s_t,a_t,s_{t+1}\right)$ is an immediate reward environment returned when the agent executes action $a_t$ at time step $t$ to make the state transit from $s_t$ to $s_{t+1}$.
Many techniques to solve MDP are divided into value-based and policy-based methods. The most popular value-based method is Q-Learning \cite{sutton2018reinforcement} which approximates the optimal Q-function $Q_*$ by $\tilde Q$ and updates its value via TD as follows,
\begin{equation} \label{uq}
    \tilde Q\left(s_t,a_t\right)\leftarrow \tilde Q\left(s_t,a_t\right) + \alpha \left(R_t + \gamma \mathop {max}\limits_{a\in \mathcal{A}}\left(s_{t+1}, a\right) - \tilde Q\left(s_t,a_t\right)\right)
\end{equation}
where $\alpha$ is the learning rate. The optimal policy $\pi_*$ is derived from greedy action, i.e., $\pi_* = \mathop{arg} \mathop{max}\limits_a Q_*\left(s,a\right)$. Mnih et al. \cite{dqn2013, dqn} propose Deep Q-Network (DQN) combining deep neural networks with Q-Learning, which minimizes the following loss function:
\begin{equation}
    \mathcal{L}\left(\theta\right) = \mathbb E_{\left<s_t,a_t,R_t,s_{t+1}\right>\sim \mathcal D}\left[\left(R_t+\gamma \mathop{max}\limits_{a\in \mathcal{A}}Q_{\theta_{-}}\left(s_{t+1},a\right)-Q_{\theta}\left(s_t, a_t\right)\right)^2\right],
\end{equation}
where $\theta$ and $\theta_-$ are parameters of Q-Network and target network fitted by a mini-batch of tuples $\left<s_t,a_t,R_t,s_{t+1}\right>$ sampled from replay buffer $\mathcal D$. 

The main idea of policy-based methods is to find optimal policy $\pi_*$ by searching policy space directly. Policy Gradient (PG) theorem \cite{sutton2018reinforcement} is as 
\begin{equation}
    \bigtriangledown_\theta \mathcal J\left(\theta\right)= \mathbb E_{s\sim \mu_{\pi_{\theta}}, a \sim \pi_\theta} \left[\bigtriangledown_\theta Q_{\pi_\theta}\left(s,a\right) \mathop{log} \pi_\theta \left(a|s\right)\right],
\end{equation}
where $\mu_{\pi_\theta}$ is the state occupancy measure under policy $\pi_\theta$. The Deterministic Policy Gradient (DPG) theorem is used in continuous action space, 
\begin{equation}
    \bigtriangledown_\theta \mathcal J\left(\theta\right) = \mathbb E_{s\sim \mu_{\pi_{\theta}}} \left[\bigtriangledown_\theta \pi_\theta \left(a|s\right) \bigtriangledown_a Q_{\pi_\theta}\left(s,a\right)|_{a=\pi_\theta \left(s\right)}\right].
\end{equation}
\subsection{Multi-agent Reinforcement Learning}
Each agent in a Multi-Agent System (MAS) solves sequential decision problems via trial-and-error contact with the environment. However, it is more complex than a single-agent scenario because the next state and reward returned by the environment are based on all agents' joint actions, making the environment non-Markovian for any agent. Stochastic Game (SG) can be used to model multi-agent sequential decision problems.
\begin{definition}[SG]
    A Stochastic game can be represented as a tuple $$\left<N, \mathcal{S}, \mathcal A^1, \cdots, \mathcal A^N, R^1, \cdots, R^N, p,\gamma \right>,$$ where $N$ is the number of agents, $\mathcal S$ is the state set of the environment, $\mathcal A^i$ is the action space of the agent $i$, $R^i: \mathcal S \times \mathcal A^1 \times \cdots \times \mathcal A^N \times \mathcal S \rightarrow \mathbb R $ is the reward function of the agent $i$, $p:\mathcal S \times \mathcal A^1 \times \cdots \times \mathcal A^N  \rightarrow \Delta\left(\mathcal S\right)$ is the transition probability based on the joint action $\bm a$ and $\gamma \in \left[0,1\right]$ is the discount factor over time.
\end{definition}
The state-action and value functions in multi-agent scenarios are defined like Eqs. (\ref{q}) and (\ref{v}), respectively. 
\begin{equation} \label{mq}
    Q_{\pi^i,\pi^{-i}}\left(s, \bm a\right) = \mathbb E_{\pi^i,\pi^{-i}}\left[\sum_{t=0}^\infty \gamma^tR^i\left(s_t,\bm a_t,s_{t+1}|\bm a_0 = \bm a, s_0 = s\right)\right],
\end{equation}
\begin{equation} \label{mv}
    V_{\pi^i,\pi^{-i}}\left(s\right) = \mathbb E_{\pi^i,\pi^{-i}}\left[\sum_{t=0}^\infty \gamma^tR^i\left(s_t,\bm a_t,s_{t+1}|s_0 = s\right)\right],
\end{equation}
where $\left(\pi^i,\pi^{-i}\right)$ is used to distinguish the policy between the agent $i$ and the other agents, similarly, we can use $\left(a^i, a^{-i}\right)$ to represent the joint action $\bm a$. The common solving SG can be divided into learning cooperation and learning communication according to whether communication between agents is involved in the execution process.
\subsubsection{Learning cooperation}
The typical approach for learning cooperation involves centralized training and decentralized execution (CTDE), utilizing global or communication information during the training process while only using the observation information of the current agent during the execution phase. It also includes value-based \cite{iql, vdn, qmix, qtran, qplex} and policy-based \cite{maddpg,Liu_Wang_Hu_Hao_Chen_Gao_2020,Ryu_Shin_Park_2020,NEURIPS2021_65b9eea6} MARL methods. 

\textbf{Value-based MARL: }
The updated rule of Eq. (\ref{uq}) is suitable for the multi-agent scenario:
\begin{equation}\label{maq}
    \tilde Q^i\left(s_t, \bm a_t \right) \leftarrow \tilde {Q}^i\left(s_t, \bm a_t \right)+\alpha \left(R^i+\mathop{max}\limits_{a^i\in\mathcal A^i} \gamma\{\tilde Q^j\left(s_t, \bm a_t\right)\}_{j\in\{1,\dots,N\}}-\tilde Q^i\left(s_t,\bm a_t\right)\right).
\end{equation}
Tampuu et al. \cite{iql} first extend the DQN to the multi-agent scenario equipping an independent DQN for each agent, i.e., only considering the agent's interaction with the environment.  The experimental results demonstrate that this fully distributed training can produce good results for simple MAS but that it is difficult to converge for complex tasks and that there is a credit assignment issue. 
Sunehag et al. \cite{vdn} overcome these issues by introducing a Value Decomposition Network (VDN) based on the CTDE. An optimal linear-valued decomposition is trained from the team reward function with VDN, and during execution, each agent uses an implicit value function based only on partial observations to make decisions. However, this decomposition is linear and can only apply to small-scale scenarios. 
Rashid et al. \cite{qmix} use an end-to-end Q-Mixing Network (QMIX) to train decentralized policies following the advantages of VND. QMIX is a complex non-linear network that constrains the joint Q-function monotonic on the Q-function of each agent. This ensures the consistency of centralized and decentralized policies and simplifies the solution for maximizing the joint action-value function in offline policy learning.
Son et al. \cite{qtran} develop an innovative MARL factorization technique called QTRAN that eliminates the structural restriction and uses a novel technique to convert the initial joint action-value function into a simple decomposition function. Although the decomposition of QTRAN is more complex computationally, it covers a broader range of MARL activities as compared to VDN and QMIX.
An approximate QTRAN performs hard in complex domains with online data collecting and requires two extra soft regularizations \cite{NEURIPS2019_f816dc0a}. As a result, effective scalability is still a challenge for cooperative MARL. 
Wang et al. \cite{qplex} use a duplex dueling network structure (QPLEX) to decompose the joint action-value function into an action-value function for each agent to address this challenge. It is made easier to learn action-value functions with a linear decomposition structure by reformulating the Individual-Global-Max (IGM) consistency as a restriction on the value range of the advantage function, which is a strong scalability value-based MARL technique.

\textbf{Policy-based MARL: }The state of the environment is determined by the action of all agents in the multi-agent scenario. The value-based method is challenging to train due to the unstable environment, and the variance of the policy-based method gets more prominent as the number of agents increases. 
Lowe et al. \cite{maddpg} proposed a variant of the actor-critic method in a multi-agent scenario - multi-agent deep deterministic policy gradient (MADDPG), which considers the action strategies of other agents in the process of reinforcement learning training for each agent, and Only individual information is considered during the testing phase. The multi-agent deterministic policy gradient can be written as 
\begin{equation}\label{maddpg}
    \bigtriangledown_{\theta^i} \mathcal J^i\left(\theta\right) = \mathbb E_{s\sim \mu_{\bm \pi_{\theta^i}}} \left[\bigtriangledown_{\theta^i} \mathop{log}\pi_{\theta^i} \left(a^i|s\right) \bigtriangledown_{a^i} Q_{\pi_{\theta^i}}\left(s,\bm a\right)|_{\bm a=\bm \pi_\theta \left(s\right)}\right].
\end{equation}
However, as the number of agents increases, the estimation error in the critic network also increases, making it difficult to scale MADDPG to larger environments. To address this limitation, researchers have proposed attention mechanisms that allow agents to focus dynamically on relevant information. For example, the MAAC \cite{maac}, G2ANet \cite{Liu_Wang_Hu_Hao_Chen_Gao_2020} and HAMA \cite{Ryu_Shin_Park_2020} algorithms use graph structures to model agent relationships and employ attention mechanisms to weigh their relevance. This approach has shown promising results in environments with a large number of agents. Another challenge in MAS is the need to adapt to changes in collaborative policies. 
The FACMAC algorithm \cite{NEURIPS2021_65b9eea6} addresses this issue by incorporating a centralized strategy gradient estimation to optimize joint action spaces. This method has been shown to outperform MADDPG and QMIX in environments with large-scale continuous actions.

\textbf{Mean-Field-based MARL: }
The above methods are all based on the CTDE training framework, effectively addressing the problem of non-Markovian environments in fully decentralized training frameworks and the problem of high computational complexity in fully centralized training frameworks. However, existing MARL methods are usually limited to a small number of agents, and scalability remains a challenging issue. Yang et al. \cite{pmlr-v80-yang18d} propose mean-field reinforcement learning (MFRL), which approximates the interaction between individuals as the interaction between individuals and the average effect of the whole group or neighboring individuals and the convergence of Nash equilibrium solutions is analyzed. Ganapathi et al. \cite{10.5555/3398761.3398813} extended MFRL to multiple types of domains and proposed the MTMFQ method. Multiple types relax a core assumption in mean-field games, which is that all agents in the environment are using almost identical strategies and have the same goals. Then they further relaxed the assumption of MFRL and extended it to partially observable domains, assuming that agents can only observe information from a fixed neighborhood or from other agents based on random distances \cite{10.5555/3463952.3464019}. Zhang et al. \cite{DBLP:conf/ijcai/ZhangY0XL21} apply mean-field theory to the value function decomposition-based MARL framework and proposed the MFVDN method, which solves the problems of homogenous agents, limited representation, and inability to execute with local information decentralized in MFRL.

\subsubsection{Learning communication}
The purpose of learning communication is for agents to learn when, with which agents, and what information to communicate, which can be categorized as reinforced and differentiable according to \cite{zhu2022survey}.

\textbf{Reinforced: } 
Foerster et al. \cite{DIAL} use DQN with a recurrent network to handle partial observability called RIAL. 
Kilinc et al. \cite{MADDPG-M} improve a DDPG algorithm enhanced by a communication medium including a concurrent learning mechanism that allows agents to decide if their private observations need to be shared with others. 
To maximize communication efficiency, Huang et al. \cite{ETCNet} propose a network named ETCNet, that uses RL to find the optimal communication protocol within bandwidth constraints. The bandwidth is minimized due to messages being sent only when necessary. 
Gupta et al. \cite{gupta2021hammer} introduce a central agent observing every observation with multiple agents only receiving local observations and no communication. The central agent determines the message each agent needs to make better decisions based on global observations, avoiding central solving of the entire problem.

\textbf{Differentiable:}
Sukhbaatar et al. \cite{NIPS2016_55b1927f} develop a neural model CommNet that lets the agents communicate continuously for fully cooperative tasks. Agents learn both their policy and communication way during training. 
To maintain effective communication, Peng et al. \cite{peng2017multiagent} propose a multi-agent Bidirectionally-Coordinated Network (BiCNet) with a vectorized actor-critic formulation. They demonstrate that BiCNet can learn advanced coordination methods without supervision.
To learn abstract representations of the interaction of agents, Jiang et al. \cite{Jiang2020Graph} propose graph convolution RL that leverages graph convolution to adapt to the underlying dynamics of the graph, with relation kernels capturing the interaction of agents.  
Wang et al.\cite{pmlr-v119-wang20i} devise a novel approach entitled IMAC, which addresses the challenges of constrained-bandwidth communication in MARL. IMAC optimizes resource usage, minimizes needless connections, and allows smooth communication protocols and schedules. It uses low-entropy messages that stick to bandwidth limits and merges the information bottleneck principle with a weight-based scheduler to produce a practical protocol.
Using an attention mechanism is insufficient as it overlooks dynamic communication and the correlation between agents' connections. To tackle this issue, Du et al. \cite{10.5555/3463952.3464010} propose a method that utilizes a normalizing flow to encode the correlation between agents' interactions, allowing for direct learning of the dynamic communication topology. This methodology proves effective in cooperative navigation and adaptive traffic control tasks.
Niu et al. \cite{10.5555/3463952.3464065} leverage a graph-attention mechanism to determine the most pertinent agent of messages and the most suitable means of delivery. 

Overall, these algorithms aim to improve the scalability and non-stationary of MAS, allowing agents to learn from the experiences of other agents and achieve better performance in complex environments.

\section{Applications of Multi-agent Reinforcement Learning} \label{Applications}
Through MARL, agents are able to learn and communicate with each other, thereby achieving more efficient task completion and better decision-making results. This method is widely used in engineering and science, for example, in smart transportation, unmanned aerial vehicles, intelligent information system, public health and intelligent medical diagnosis, smart manufacturing, financial trade, network security, smart education, and RL for science. 

\subsection{Smart Transportation}
\begin{table}[]
\caption{Correspondence between smart transportation and RL methods.}
\label{transportation}
\centering
\begin{tabularx}{\textwidth}{|XX|c|c|c|}
\hline
\multicolumn{2}{|c|}{Applications}                                                                   & Papers                             & Methods                          & SA/MA \\ \hline
\multicolumn{1}{|X|}{\multirow{8}{=}{\centering Smart Transportation}} & \multirow{3}{=}{\centering Traffic light control} & \cite{7508798}                     & DQN-based \cite{dqn}             & SA       \\ \cline{3-5} 
\multicolumn{1}{|X|}{}                                      &                                        & \cite{9103316}                     & MADDPG-based \cite{maddpg}       & MA       \\ \cline{3-5} 
\multicolumn{1}{|X|}{}                                      &                                        & \cite{9681232}                     & Game theoretic                   & MA       \\ \cline{2-5} 
\multicolumn{1}{|X|}{}                                      & \multirow{5}{=}{\centering Auto-driving}          & \cite{8638814}                     & Dynamic coordination graph       & MA       \\ \cline{3-5} 
\multicolumn{1}{|X|}{}                                      &                                        & \cite{pmlr-v155-zhou21a}           & Auto-driving simulation platform & MA       \\ \cline{3-5} 
\multicolumn{1}{|X|}{}                                      &                                        & \cite{10.1007/978-3-030-47358-7_7} & DQN-based \cite{dqn}             & MA       \\ \cline{3-5} 
\multicolumn{1}{|X|}{}                                      &                                        & \cite{9694460}                     & AC-based \cite{ac}     & SA       \\ \cline{3-5} 
\multicolumn{1}{|X|}{}                                      &                                        & \cite{zhou2022multi}               & AC-based \cite{a3c}    & MA       \\ \hline
\end{tabularx}
\end{table}


Smart transportation makes use of advanced technologies like the Internet of Things (IoT) and AI to increase safety, improve transportation efficiency, and reduce its negative environmental effects.
In MARL-based smart transportation, we describe two known scenarios: traffic light control and auto-driving  and present the role of humans in these intelligent systems. The correspondence between this application and RL methods is shown in Table \ref{transportation}.

\textbf{Traffic light control:} Li et al. \cite{7508798} use DQN to obtain the optimal policy in sight of the variety of the control action and the state and demonstrate the potential of DRL in traffic light control. However, the control of traffic lights needs to consider the situation of multiple intersections. Wu et al. \cite{9103316} combine MADDPG with Long-short-term Memory (LSTM) for multi-intersection traffic light control coordination. The use of LSTM is appropriate to address the environmental instability forced on by partial observable states. They take into account both the cars and the pedestrians waiting to cross the street. Zhu et al. \cite{9681232} propose a Bi-hierarchical Game-theoretic (BHGT) to solve network-wide traffic signal control problems. They evaluate the state of the network-wide traffic based on the collection data of trips. The experiment shows that BHGT efficiently reduces the network-wide travel delay. 

\textbf{Auto-driving:} Chao et al. \cite{8638814}  simulate the dynamic topography during vehicle interactions using a dynamic coordination graph and put forward two fundamental learning strategies to coordinate the driving actions for a fleet of vehicles. Additionally, they propose a number of extension mechanisms in order to adapt to the complex scenario with any number of vehicles.
Zhou et al. \cite{pmlr-v155-zhou21a} build an autonomous driving simulation platform to realize more realistic and diverse interactions.
Bhalla et al. \cite{10.1007/978-3-030-47358-7_7} propose two novel centralized training based on DQN and a memory block to execute decentralized, which achieve better cumulative reward in autonomous driving.
Huang et al. \cite{9694460} propose a sample efficient DRL framework including imitative expert priors and RL. The agent learns expert policy from the prior human knowledge and is guided by minimizing the KL divergence between the policy of the agent and the imitative expert. 
Zhou et al. \cite{zhou2022multi} propose a MARL framework composed of a brand-new local reward and scheme for sharing parameters for lane-changing decision makings.

As a system that involves both physical and digital components, it requires the active participation and cooperation of humans to achieve its full potential. Humans play a crucial role in the operation and management of systems for transportation, from designing and building infrastructure to using and maintaining vehicles to making decisions about routing and scheduling. Thus, the success of smart transportation ultimately depends on how well it can integrate and leverage the capabilities of both humans and machines in a seamless and effective manner. However, the current state of research on MARL-based smart transportation is without adequately address the decision priority between human control and intelligent algorithms. Given the continuously evolving nature of both human behavior and city traffic, situations such as traffic accidents and surges in vehicles can make it challenging to manage traffic jams solely through traffic signal control. In such scenarios, human intervention becomes necessary. Similarly, in instances where self-driving cars encounter hazardous situations that were not anticipated during training, relinquishing control to the human driver is critical. Defining the optimal decision priority between humans and agents remains an unresolved issue.



\subsection{Unmanned Aerial Vehicles}
\begin{table}[]
\centering
\caption{Correspondence between unmanned aerial vehicles and RL methods.}
\label{uav}
\begin{tabularx}{\textwidth}{|XX|c|c|c|}
\hline
\multicolumn{2}{|c|}{Applications}                                                                                & Papers                                                           & Methods                    & SA/MA \\ \hline
\multicolumn{1}{|X|}{\multirow{9}{=}{\centering Unmanned Aerial Vehicles}} & \multirow{3}{=}{\centering Cluster control}                & \cite{maciel2019online}                                          & DQN-based \cite{dqn}       & SA       \\ \cline{3-5} 
\multicolumn{1}{|X|}{}                                          &                                                 & \cite{9001167}                                                   & DDPG-based \cite{ddpg}     & MA       \\ \cline{3-5} 
\multicolumn{1}{|X|}{}                                          &                                                 & \cite{9209079}                                                   & MADDPG-based \cite{maddpg} & MA       \\ \cline{2-5} 
\multicolumn{1}{|X|}{}                                          & \multirow{4}{=}{\centering Environmental   monitoring}     & \cite{journals/corr/abs-1803-07250}                              & DQN-based \cite{dqn}       & MA       \\ \cline{3-5} 
\multicolumn{1}{|X|}{}                                          &                                                 & \cite{julian2019distributed}                                     & DQN-based \cite{dqn}       & MA       \\ \cline{3-5} 
\multicolumn{1}{|X|}{}                                          &                                                 & \cite{9172262}                                                   & TRPO-based \cite{trpo}     & MA       \\ \cline{3-5} 
\multicolumn{1}{|X|}{}                                          &                                                 & \cite{9453825}                                                   & DQN-based \cite{dqn}       & MA       \\ \cline{2-5} 
\multicolumn{1}{|X|}{}                                          & \multirow{2}{=}{\centering Collaborative   transportation} & \cite{10.1007/978-981-19-2635-8_71,en15197426} & MAAC-based \cite{maac}     & MA       \\ \cline{3-5} 
\multicolumn{1}{|c|}{}                                          &                                                 & \cite{9993797}                                                   & MADDPG-based \cite{maddpg} & MA       \\ \hline
\end{tabularx}
\end{table}
In MARL-based Unmanned Aerial Vehicles (UAVs) applications, we describe three known scenarios: cluster control \cite{maciel2019online,9001167,9209079,XU2020196,QIU2020515,xu_chen_2022,Xu2022}, environmental monitoring  \cite{journals/corr/abs-1803-07250,julian2019distributed,9172262, 9453825}, and collaborative transportation \cite{10.1007/978-981-19-2635-8_71,en15197426,9993797}. 
The correspondence between this application and RL methods is shown in Table \ref{uav}.

\textbf{Cluster control:} Maciel-Pearson et al. \cite{maciel2019online} make use of DRL to improve the ability of UAVs to automatically navigate when the environments are various. The approach uses a double state-input strategy that combines positional information with feature maps from the current scene. This approach is tested and shown to outperform other DQN variants and has the ability to navigate through multiple unknown environments and extreme weather conditions. A two-stage RL method is proposed by Wang et al. \cite{9001167} for multi-UAV collision avoidance to address the issues of high variance and low reproducibility, where supervised training is in the first stage and policy gradient is in the next stage. Wang et al. \cite{9209079} propose a trajectory control method according to MARL, which introduced a low-complexity approach to optimize the offloading decisions of the user equipment given the trajectories of UAVs. The results show that the proposed approach has promising performance. 

\textbf{Environmental monitoring:} Pham et al. \cite{journals/corr/abs-1803-07250} propose a distributed MARL algorithm to achieve complete coverage of an unfamiliar area while minimizing overlapping fields of view.  Julian and Kochenderfer \cite{julian2019distributed}
present two DRL approaches for controlling teams of UAVs to monitor wildfires. The approaches accommodate the problem with uncertainty and high dimensionality and allow the UAV to accurately track the wildfire expansions and outperform existing controllers. The approaches scale with different numbers of UAVs and generalize to various wildfire shapes.
Walker et al. \cite{9172262} propose a method for indoor target-finding by combining Partially Observable MDP (POMDP) and DRL. The framework consists of two stages: planning and control. Global planning is done using an online POMDP solver, while local control is done using Deep RL. Mou et al. \cite{9453825} propose a hierarchical UAV swarm architecture based on the DRL algorithm for solving the 3D irregular terrain surface coverage problem. A geometric approach is used to divide the 3D terrain surface into weighted 2D patches. A coverage trajectory algorithm is designed for low-level follower UAVs to achieve specific coverage tasks within patches. For high-level leader UAVs, a swarm DQN algorithm is proposed to choose patches, which integrates Convolutional Neural Networks (CNNs) and mean embedding methods to address communication limitations. 

\textbf{Collaborative transportation: } Jeon et al. \cite{10.1007/978-981-19-2635-8_71} design a UAV logistics delivery service environment using Unity to evaluate MADRL-based models, and Jo \cite{en15197426} propose a fusion-multi-actor-attention-critic (F-MAAC) model based on the MAAC. It is shown from the results that F-MAAC outperformed MAAC in terms of the total number of deliveries completed during a specific period and the total number of deliveries completed over the same distance.
Our previous work \cite{9993797} develops a virtual platform for multi-UAVs collaborative transport using AirSim \cite{10.1007/978-3-319-67361-5_40} and proposed recurrent-MADDPG with domain randomization technique to achieve MARL sim2real transfer.

By utilizing MARL, UAV systems can make autonomous decisions and collaborations in various scenarios, leading to more efficient task completion. However, existing works do not consider the command and interaction between ground workstations and operators for UAV systems, and the robustness and safety of MARL are deficient. When a UAV encounters interference and cannot make the correct decisions, it can cause serious harm to human society. Considering the interaction between intelligent UAV systems and humans to achieve more efficient and safer UAV systems is one of the goals in future 10-20 years.

\subsection{Intelligent Information System}
\begin{table}[]
\centering
\caption{Correspondence between intelligent information system and RL methods.}
\label{information}
\begin{tabularx}{\textwidth}{|XX|c|c|c|}
\hline
\multicolumn{2}{|c|}{Applications}                                                                                      & Papers                                                          & Methods                                                  & SA/MA \\ \hline
\multicolumn{1}{|X|}{\multirow{12}{=}{\centering Intelligent Information System}} & \multirow{6}{=}{\centering Natural language   processing} & \cite{li2016deep}                                               & REINFORCE-based \cite{REINFORCE}                         & SA       \\ \cline{3-5} 
\multicolumn{1}{|X|}{}                                                 &                                                & \cite{9025776}                                                  & AC-based \cite{ac}                                       & SA       \\ \cline{3-5} 
\multicolumn{1}{|X|}{}                                                 &                                                & \cite{Lu_Zhang_Chen_2019}                                       & DQN-based \cite{dqn}                                     & SA       \\ \cline{3-5} 
\multicolumn{1}{|X|}{}                                                 &                                                & \cite{8801910}                                                  & REINFORCE,AC,DQN \cite{REINFORCE,ac,dqn} & SA       \\ \cline{3-5} 
\multicolumn{1}{|X|}{}                                                 &                                                & \cite{li2017paraphrase}                                         & AC-based \cite{ac}                                       & SA       \\ \cline{3-5} 
\multicolumn{1}{|X|}{}                                                 &                                                & \cite{NEURIPS2022_b1efde53}                                     & PPO-based \cite{ppo}                                     & SA       \\ \cline{2-5} 
\multicolumn{1}{|X|}{}                                                 & \multirow{3}{=}{\centering Programming   generation}      & \cite{NEURIPS2022_8636419d}                                     & REINFORCE \cite{REINFORCE}                               & SA       \\ \cline{3-5} 
\multicolumn{1}{|X|}{}                                                 &                                                & \cite{shojaee2023execution}                                     & PPO-based \cite{ppo}                                     & SA       \\ \cline{3-5} 
\multicolumn{1}{|X|}{}                                                 &                                                & \cite{ESNAASHARI2021115446}                                     & DQN-based \cite{dqn}                                     & SA       \\ \cline{2-5} 
\multicolumn{1}{|X|}{}                                                 & \multirow{3}{=}{\centering Recommender system}            & \cite{10.1145/3383313.3412233,10.1145/3269206.3272021,10016386} & MADDPG-based \cite{maddpg}                               & MA       \\ \cline{3-5} 
\multicolumn{1}{|X|}{}                                                 &                                                & \cite{10.1145/3109859.3109914}                                  & Learning communication                                   & MA       \\ \cline{3-5} 
\multicolumn{1}{|X|}{}                                                 &                                                & \cite{10.1145/3331184.3331237}                                  & IQL-based \cite{iql}                                     & MA       \\ \hline
\end{tabularx}
\end{table}
MARL has tremendous potential for applications in intelligent information systems, including natural language processing (NLP) \cite{Uc-Cetina2023, li2016deep, 9025776, Lu_Zhang_Chen_2019, chen-etal-2017-line, SU201824, 8801910, li2017paraphrase}, programming generation \cite{li2017paraphrase, shojaee2023execution, ESNAASHARI2021115446}, and recommender systems \cite{10.1145/3383313.3412233, 10.1145/3109859.3109914, 10.1145/3331184.3331237, 10.1145/3269206.3272021, 10016386}. Techniques based on SARL have been studied in NLP and programming generation, and we will summarize these studies and point out the significant advantages of MARL in these applications. The correspondence between this application and RL methods is shown in Table \ref{information}.

\textbf{Natural language processing: }
Li et al. \cite{li2016deep} describe how RL can be applied to chatbot dialogue generation to predict the impact of current actions on future rewards. By utilizing a policy gradient approach to optimize long-term rewards defined by the developer, the model learns all possible strategies for speaking in an infinite action space, resulting in more interactive and consistent conversation generation for chatbots. 
Yang et al. \cite{9025776} combine multi-task learning and RL to present a personalized dialog system called MRPDG. Three kinds of rewards are used to guide the model to produce highly rewarded dialogs.
In order to address the problems of sparse rewards and few successful dialogues, Lu et al. \cite{Lu_Zhang_Chen_2019} propose two complex methods for hindsight experience replay. 
During the RL training process, chatbot agents can be made to generate more authentic dialogues by introducing human-relevant evaluation metrics. Chen et al. \cite{chen-etal-2017-line} present a framework called "companion teaching" in which a human teacher guides the machine in real-time during the learning process and uses example actions of the teacher to improve policy learning. Su et al. \cite{SU201824} present two approaches to address the challenge of measuring rewards in real-world dialogue system applications.
Keneshloo et al. \cite{8801910} use RL to solve the effects of sequence-to-sequence exposure bias and inconsistency between training and test measurements.
Li et al. \cite{li2017paraphrase} propose a new framework that includes a generator and an evaluator for learning from data. The generator is a learning model for paragraph generation, and the evaluator is a matching model used to provide a reward function for RL.
Large language models may produce fake or useless outputs. Ouyang et al. \cite{NEURIPS2022_b1efde53} introduce RL with human feedback to fine-tune GPT-3 to reduce unwanted outputs and propose a language model called instructGPT. They believe that it is important to use human feedback to make the output of large language models close to human intention. 

\textbf{Programming generation: }
Le et al. \cite{NEURIPS2022_8636419d} design a program synthesis framework called CodeRL that uses pre-trained language models and RL to generate programs.
Shojaee et al. \cite{shojaee2023execution} integrate a pre-trained programming language model with PPO to optimize the model through execution feedback and present a new code generation framework called PPOCoder. 
Software testing is essential for quality assurance but expensive and time-consuming. Esnaashari et al. \cite{ESNAASHARI2021115446} propose a new method using a memetic algorithm with RL as a local search that outperforms traditional evolutionary or heuristic algorithms in speed, coverage, and evaluation.

MARL has advantages over SARL in  NLP and programming generation due to its stronger collaboration ability and adaptability. In NLP, MARL can be used for tasks such as chatbots and text translation. In these tasks, multiple agents can work together to learn the knowledge and skills of a conversational system, thereby improving its performance and interaction experience. For programming generation, MARL is usually more suitable for scenarios that require the generation of complex systems or large-scale software. This is because in MARL, each agent can be responsible for generating a part of the code, and the whole system can be built through collaboration. This approach can improve the efficiency and quality of the generated code and can reduce the repetition and error rate of the code.

\textbf{Recommender system: }
He et al. \cite{10.1145/3383313.3412233}  propose a MARL method with communication restrictions to address sub-optimal global strategies due to the lack of cooperation among optimization teams.  
Zhang et al. \cite{10.1145/3109859.3109914} propose a novel dynamic, collaborative recommendation method utilizing MARL for recommending academic collaborators, optimizing collaborator selection from different similarity measures.
To improve communication efficiency on Twitter-like social networking, Gui et al. \cite{10.1145/3331184.3331237} propose a MARL by combining dozens of more historical tweets to choose a set of users.
Jin et al. \cite{10.1145/3269206.3272021} propose a method for optimizing bids using MARL to achieve specific goals, such as maximizing revenue and return on investment for real-time advertising. The method uses a clustering approach to assign strategic bidding agents to each advertiser cluster and proposes a practical distributed coordinated multi-agent bidding to balance competition and cooperation among advertisers.
Li and Tong \cite{10016386} propose a social MARL framework named MATR, where one agent captures the dynamic preferences of users while the other exploits social networks to reduce data sparsity and cold starts. The state representation module aims to learn from social networks and user rating matrices, using trust inference and feature aggregation modeling to optimize the use of social networks.

MARL has many advantages in intelligent information processing, but the lack of robustness and transparency prevents MARL decisions from being trusted by humans. In order to apply MARL to the real world, it is first necessary to improve its trustworthiness, and in addition, RL with human feedback needs to be further considered to make the generated language more realistic, the programming more efficient, and the recommended content more attractive.

\subsection{Public Health and Intelligent Medical Diagnosis}

\begin{table}[]
\centering
\caption{Correspondence between public health and intelligent medical diagnosis and RL methods.}
\label{health}
\begin{tabularx}{\textwidth}{|XX|c|c|c|}
\hline
\multicolumn{2}{|c|}{Applications}                                                                                                          & Papers                                                                                                                   & Methods                & SA/MA \\ \hline
\multicolumn{1}{|X|}{\multirow{8}{=}{\centering Public Health and Intelligent Medical Diagnosis}} & \multirow{2}{=}{\centering COVID-19} & \cite{9551174,   Khalilpourazari2022,10.3389/fpubh.2021.744100}                                                          & DQN-based \cite{dqn}   & SA       \\ \cline{3-5} 
\multicolumn{1}{|X|}{}                                                                 &                                                    & \cite{Zheng2021}                                                                                                         & DDPG-based \cite{ddpg} & SA       \\ \cline{2-5} 
\multicolumn{1}{|X|}{}                                                                 & \multirow{4}{=}{\centering Medical image processing}          & \cite{JALALI2021107675,9855449}                                                                                          & DQN-based \cite{dqn}   & SA       \\ \cline{3-5} 
\multicolumn{1}{|X|}{}                                                                 &                                                    & \cite{jpm12020309,zheng2021multi,10.1007/978-3-030-32251-9_29,10.1007/978-3-030-66843-3_18,10.1007/978-3-030-32251-9_29} & DQN-based \cite{dqn}   & MA       \\ \cline{3-5} 
\multicolumn{1}{|X|}{}                                                                 &                                                    & \cite{Liao_2020_CVPR,9311659}                                                                                            & A3C-based \cite{a3c}   & MA       \\ \cline{3-5} 
\multicolumn{1}{|X|}{}                                                                 &                                                    & \cite{10.1007/978-3-030-78191-0_59}                                                                                      & AC-based \cite{ac}     & MA       \\ \cline{2-5} 
\multicolumn{1}{|X|}{}                                                                 & \multirow{2}{=}{\centering Disease diagnosis}                 & \cite{pmlr-v68-ling17a,   ling-etal-2017-learning,tang2016inquire}                                                       & DQN-based \cite{dqn}   & SA       \\ \cline{3-5} 
\multicolumn{1}{|X|}{}                                                                 &                                                    & \cite{10010747}                                                                                                          & DQN-based \cite{dqn}   & MA       \\ \hline
\end{tabularx}
\end{table}

MARL is widely explored and applied in public health and intelligent medical diagnosis. For example, MARL can be applied in COVID-19 prediction and management, medical image processing, and disease diagnosis to improve disease prevention, diagnosis, and treatment efficiency and accuracy. The correspondence between this application and RL methods is shown in Table \ref{health}.

\textbf{COVID-19 prediction and diagnosis: }
Khalilpourazari et al. \cite{9551174, Khalilpourazari2022} present the Hybrid Q-learning-based algorithm (HQLA) as a solution to predict the COVID-19 pandemic. HQLA accurately reflects the future trend in France and Quebec, Canada. Furthermore, their analysis also provides critical insights into pandemic growth and factors that policymakers should consider when making social measures. 
Kumar et al. \cite{10.3389/fpubh.2021.744100} utilize two learning algorithms, DL and RL, to forecast COVID-19, where LSTM is used to forecasts newly affected individuals, losses, and cures in the coming days, and DQN is suggested for optimizing predictive outcomes based on symptoms.
Zheng et al. \cite{Zheng2021} propose developing MDPs to model the oxygen flow trajectory and health outcomes of COVID-19 patients. Using Deep Deterministic Policy Gradient (DDPG), an optimal oxygen control policy is obtained for each patient, resulting in a reduced mortality rate.

Regarding the prediction and diagnosis of COVID-19, existing studies are based on SARL. compared with SARL, MARL can be responsible for different tasks, such as virus transmission model prediction and clinical diagnosis, separately and then complete the task through communication and collaboration. In addition, the COVID-19 epidemic develops rapidly and is influenced by multiple factors, and MARL can better handle the uncertainty and complexity. Therefore, we believe that MARL has excellent potential in this area.

\textbf{Medical image processing: }
X-ray images have become crucial for expediting the diagnostics of COVID-19. Jalali et al. \cite{JALALI2021107675} propose an ensemble of CNNs to differentiate COVID-19 patients from non-patients according to an automated X-ray image. The selective ensemble approach utilizes DQN to heighten model accuracy while reducing the required classifiers.
Chen et al. \cite{9855449} suggest an RL-based detection framework to quickly and effectively diagnose COVID-19. They build a mixed loss, enabling efficient detection of the virus. Additionally, they propose a prediction framework that allows for integrating multiple detection frameworks through parameter sharing. This allows for the prediction of disease progression without the need for additional training.
Allioui et al. \cite{jpm12020309} develop a new method for more efficient automatic image segmentation that employs MARL. This approach addresses mask extraction difficulties and uses a modified version of the DQN to identify masks in CT images of COVID-19 patients. 
MARL can be used for interactive image segmentation, where each voxel is an agent with a shared behavior policy to reduce exploration space and dependence among voxels. \cite{Liao_2020_CVPR} is for the field of medical image segmentation, considering clinical criteria, using MARL to solve the problem, reducing the exploration space, and using a sharing strategy to capture the dependencies between pixels;
While \cite{9311659} is for interactive image segmentation, using MDP and MARL models to model iterative segmentation, introducing a boundary-based reward function to update the segmentation strategy.
Zheng et al. \cite{zheng2021multi} use a MARL approach to prostate localization in Magnetic Resonance (MR) images. They create a communication environment by sharing convolutions and maintaining independent action policy via distinct fully connected layers for each agent.
Anatomical landmark detection is crucial in medical image analysis. Vlontzos et al.\cite{10.1007/978-3-030-32251-9_29} present a novel approach using MARL to detect multiple landmarks simultaneously. This theory suggests that the positioning of anatomical landmarks in human anatomy is interdependent and not random. It can accommodate $K$ agents to detect $K$ different landmarks with implicit inter-communication.
Leroy et al. \cite{10.1007/978-3-030-66843-3_18} develop a communicative MARL framework, aiding in detecting landmarks in MR images. In contrast to \cite{10.1007/978-3-030-32251-9_29}, agent communication is explicit.
Kasseroller et al. \cite{10.1007/978-3-030-78191-0_59} propose a solution to the long inference time caused by DQN-based methods being limited to a discrete action space. They recommend using a continuous action space to allow the agent to move smoothly in any direction with varying step sizes, resulting in fewer required steps and increased landmark identification accuracy.

\textbf{Disease diagnosis: }
Ling et al. \cite{pmlr-v68-ling17a, ling-etal-2017-learning} propose an RL-based method to improve clinical diagnostic inferencing. This approach can extract clinical concepts, integrate external evidence, and identify accurate diagnoses, which is especially beneficial in cases with limited annotated data. The system uses a DQN architecture and a reward function to optimize accuracy during training.
Tang et al. \cite{tang2016inquire} introduce a new neural symptom checker that employs an ensemble model. They incorporate an RL framework to develop inquiry and diagnosis policies as MDPs without using previous approximation methods. Furthermore, they develop a model for each anatomical section reflective of the practices of various hospital departments. This new approach offers improved user experience and significant enhancements in disease prediction accuracy over current models.
Rajesh et al. \cite{10010747} created the IMRLDPTR system, which uses mobile agents to collect data from multiple sources and generates rule sets for different disease categories.

MARL has many benefits in public health and intelligent medical diagnosis, such as the ability to handle highly complex tasks and to consider the interaction of multiple factors and variables. However, MARL also has some drawbacks, such as low transparency of the learning process and decision results, making it difficult to understand the decision process and behavior of the model. In addition, the robustness of MARL is poor, and the decisions are sensitive to perturbations. Therefore, the above drawbacks must be addressed when applying MARL to this field.

\subsection{Smart Manufacturing}
Smart manufacturing is the integration of advanced technologies, e.g., IoT, AI, and so on, into the manufacturing process to optimize the production process. As for smart manufacturing, MARL is a promising approach. In the context of smart manufacturing, MARL can be utilized as a tool for production scheduling, shop industrial robot control, quality control, and equipment maintenance to achieve an intelligent and efficient production process \cite{LI202375}. The correspondence between this application and RL methods is shown in Table \ref{manufacturing}.

\begin{table}[]
\centering
\caption{Correspondence between smart manufacturing and RL methods.}
\label{manufacturing}
\begin{tabularx}{\textwidth}{|XX|c|c|c|}
\hline
\multicolumn{2}{|c|}{Applications}                                                                    & Papers                                          & Methods                    & SA/MA \\ \hline
\multicolumn{1}{|X|}{\multirow{12}{=}{\centering Smart Manufacturing}} & \multirow{6}{=}{\centering Job shop scheduling}    & \cite{WANG2022102324}                           & QMIX-based \cite{qmix}     & MA       \\ \cline{3-5} 
\multicolumn{1}{|X|}{}                                      &                                         & \cite{ZHANG2022102412}                           & PPO-based \cite{ppo}       & MA       \\ \cline{3-5} 
\multicolumn{1}{|X|}{}                                      &                                         & \cite{Jing2022}                                 & DDPG-based \cite{ddpg}     & MA       \\ \cline{3-5} 
\multicolumn{1}{|X|}{}                                      &                                         & \cite{POPPER202263,9590925}                     & PPO-based \cite{ppo}       & MA       \\ \cline{3-5} 
\multicolumn{1}{|X|}{}                                      &                                         & \cite{ZHANG2023110083}                          & DQN-based   \cite{dqn}     & MA       \\ \cline{3-5} 
\multicolumn{1}{|X|}{}                                      &                                         & \cite{10.1007/978-3-030-41913-4_1}              & IQL-based \cite{iql}       & MA       \\ \cline{2-5} 
\multicolumn{1}{|X|}{}                                      & \multirow{4}{=}{\centering Industrial robots}      & \cite{agrawal_won_sharma_deshpande_mccomb_2021} & PPO-based \cite{ppo}       & MA       \\ \cline{3-5} 
\multicolumn{1}{|X|}{}                                      &                                         & \cite{Tan2019}                                  & DQN-based   \cite{dqn}     & MA       \\ \cline{3-5} 
\multicolumn{1}{|X|}{}                                      &                                         & \cite{krnjaic2022scalable}                      & MADDPG-based \cite{maddpg} & MA       \\ \cline{3-5} 
\multicolumn{1}{|X|}{}                                      &                                         &  \cite{9376433}                       & AC-based \cite{ac}         & MA       \\ \cline{2-5} 
\multicolumn{1}{|X|}{}                                      & \multirow{2}{=}{\centering Preventive maintenance} & \cite{SU2022116323}                             & MADDPG-based \cite{maddpg} & MA       \\ \cline{3-5} 
\multicolumn{1}{|X|}{}                                      &                                         & \cite{RUIZRODRIGUEZ2022102406}                  & PPO-based \cite{ppo}       & MA       \\ \hline
\end{tabularx}
\end{table}

\textbf{Job shop scheduling} is a key challenge in smart manufacturing because it involves complex decision-making processes and resource allocation problems. Traditional approaches are usually based on rules or static algorithms, but these approaches frequently fall short of adjusting to the changing production environment. In recent years, MARL has been introduced to job shop scheduling to improve the efficiency and accuracy of shop floor task scheduling by learning and adapting strategies from a progressively changing environment.
In the resource preemption that addresses the high-dimensional action space problem. A MARL algorithm for job scheduling is proposed in \cite{WANG2022102324}. In the algorithm, the environment is modeled as a Markov decision process which is decentralized and partially observable. And every job is regarded as an agent which selects the available robot. 
Zhang et al. \cite{ZHANG2022102412} propose a multi-agent manufacturing system for efficient and autonomous personalized order processing in a changeable workshop environment. The manufacturing equipment is built as an agent with an AI scheduler, which generates excellent production strategies in sight of the workshop state and is periodically trained through the PPO algorithm \cite{ppo}. This algorithm can tackle resource or task disturbances and obtain solutions that satisfy different performance metrics.
Jing et al. \cite{Jing2022} address the flexible job shop scheduling issues by utilizing a graph-based MARL with centralized learning decentralized execution. The approach uses a directed acyclic graph to simulate the flexible job shop scheduling issues and predicts the connection probability among edges to adjust the scheduling strategy.
Popper et al. \cite{POPPER202263,9590925} use MARL to deal with the issues of flexible job shop scheduling with multiple objectives.
Zhang et al. \cite{ZHANG2023110083} propose a new model called DeepMAG for flexible job shop scheduling according to MARL. DeepMAG provides each machine and job with an agent, and they work together to find the best action.
In Industry 4.0, a user-friendly MARL tool for the job shop scheduling problem is designed in \cite{10.1007/978-3-030-41913-4_1}, which provides users with the chance to communicate with the learning algorithms. Users can either maintain the optimal schedule produced by Q-Learning or change it to meet constraints.

\textbf{Industrial robots} have a growing amount of influence on industrial manufacturing. However, with the increasing complexity of production tasks, it is often difficult for individual robots to complete tasks effectively. MARL is widely used in smart manufacturing robots. 
Agrawal et al. \cite{agrawal_won_sharma_deshpande_mccomb_2021} propose a framework based on MARL that integrates job scheduling and navigation control for an autonomous mobile robot-operated shop floor. 
To address the challenge of increasing demands for customization and rapid product iterations, Tan et al. \cite{Tan2019} propose a multi-agent model for the industrial robot assembly process, and the communication of agents which have real-time data acquisition and fusion is studied. Besides, they also propose an excellent algorithm for planning and scheduling industrial robot assembly using a MARL approach. 
Krnjaic et al. \cite{krnjaic2022scalable} use MARL to optimize order-picking systems in commercial warehouses. The goal is to improve efficiency and flexibility while minimizing resource constraints. The MARL framework is applicable to various configurations of warehouses and allows agents to learn how to cooperate optimally with one another. 
Lan et al. \cite{9376433} explore the use of MARL to optimize coordination in a multi-robot pick-and-place system for smart manufacturing. 

\textbf{Preventive maintenance: } With the increasing scale and productivity of the manufacturing industry, how to design useful preventive maintenance strategies to guarantee the steady operation of production systems has become a vital issue in the manufacturing field. The MARL approach has provided a new idea to address this issue. 
Due to the problem of action space explosion, traditional RL methods are difficult to be applied directly. Therefore, \cite{SU2022116323} adopts a MARL-based approach in a manufacturing system to model every machine as a collaborative intelligence and implements adaptive learning through the multi-agent value decomposition Actor-Critic algorithm to obtain an efficient and cost-reasonable preventive maintenance strategy.
\cite{RUIZRODRIGUEZ2022102406} present a multi-agent approach using RL to coordinate maintenance scheduling and dynamically assign tasks to technicians with various skills under the uncertainty of multiple machine failures.
%

MARL shows potential applications in smart manufacturing and achieves some stunning results. However, this approach has challenges in scalability and is difficult to scale to situations with a high number of agents. It also suffers from poor generalization, which makes it difficult to be applied well to real scenarios. In addition, smart manufacturing is a task that involves human-computer interaction, so human behavior and human-computer priority switching need to be considered when applying MARL. All these factors need to be fully considered when designing and implementing MARL algorithms to ensure the reliability and applicability of the models.

\subsection{Financial Trade}
Financial trading is a challenging activity that requires fast judgment and adjustment to continuously changing market conditions. Single-agent approaches and DL techniques from the past are no longer adequate to meet market expectations. MARL offers a fresh idea for tackling the difficulties in financial trade by combining collaboration and competition among various agents. We summarize the applications of MARL in financial trade from the perspectives of portfolio management \cite{Pham2021, lee2020maps, huang2022mspm, Ma2023, SHAVANDI2022118124}, trading strategy optimization \cite{9931995,qiu2021multi,patel2018optimizing,10.1145/3383455.3422570}, and risk management \cite{BAJO20126921, ganesh2019reinforcement, HE2023109985}. The correspondence between this application and RL methods is shown in Table \ref{trade}.

\begin{table}[]
\centering
\caption{Correspondence between financial trade and RL methods.}
\label{trade}
\begin{tabularx}{\textwidth}{|XX|c|c|c|}
\hline
\multicolumn{2}{|c|}{Applications}                                                                       & Papers                                              & Methods                            & SA/MA \\ \hline
\multicolumn{1}{|X|}{\multirow{10}{=}{\centering Financial Trade}} & \multirow{2}{=}{\centering Portfolio management}          & \cite{Pham2021,Ma2023}                              & MADDPG-based \cite{maddpg}         & MA    \\ \cline{3-5} 
\multicolumn{1}{|X|}{}                                  &                                                & \cite{lee2020maps,huang2022mspm,SHAVANDI2022118124} & DQN-based \cite{dqn}               & MA    \\ \cline{2-5} 
\multicolumn{1}{|X|}{}                                  & \multirow{5}{=}{\centering Trading strategy optimization} & \cite{9931995}                                      & MFRL-based \cite{pmlr-v80-yang18d} & MA    \\ \cline{3-5} 
\multicolumn{1}{|X|}{}                                  &                                                & \cite{qiu2021multi}                                 & MADDPG-based \cite{maddpg}         & MA    \\ \cline{3-5} 
\multicolumn{1}{|X|}{}                                  &                                                & \cite{patel2018optimizing}                          & DQN-based \cite{dqn}               & MA    \\ \cline{3-5} 
\multicolumn{1}{|X|}{}                                  &                                                & \cite{10.1145/3383455.3422570}                      & Double-Q-based \cite{doubleq}      & MA    \\ \cline{3-5} 
\multicolumn{1}{|X|}{}                                  &                                                & \cite{bao2019fairness}                              & DDPG-based \cite{ddpg}             & MA    \\ \cline{2-5} 
\multicolumn{1}{|X|}{}                                  & \multirow{3}{=}{\centering Risk management}               &  \cite{BAJO20126921}                     & Multi-agent System                 & MA    \\ \cline{3-5} 
\multicolumn{1}{|X|}{}                                  &                                                & \cite{ganesh2019reinforcement}                      & PPO-based \cite{ppo}               & MA    \\ \cline{3-5} 
\multicolumn{1}{|X|}{}                                  &                                                & \cite{HE2023109985}                                 & DQN-based \cite{dqn}               & MA    \\ \hline
\end{tabularx}
\end{table}

\textbf{Portfolio management:} 
In portfolio management, MARL can help investors better optimize asset allocation and improve returns. Multiple agents make investment decisions and are trained to achieve optimal investment portfolios and returns.
For a portfolio of 10 equities on the Vietnam stock market, Pham et al. \cite{Pham2021} use MARL to create an automatic hedging strategy. They develop a simulator including transaction fees, taxes, and settlement dates for training the RL agent. The agent can get knowledge of trading and hedging to minimize losses and maximize earnings. It also protected portfolios and generated positive profits in case of a systematic market collapse.
Lee et al. \cite{lee2020maps} propose a new investment strategy called a MARL-based portfolio management system (MAPS) that uses a cooperative system of independent "investor" agents to create a diversified portfolio. The agents are trained to act in a variety of ways and maximize their return using a thought-out loss function. 
To address the scalability and re-usability in RL-based portfolio management, Huang and Tanaka \cite{huang2022mspm} propose a MARL-based system with Evolving Agent Module (EAM) and the Strategic Agent Module (SAM). EAM generates signal-comprised information for a particular asset using a DQN agent. In contrast, SAM uses a PPO agent for portfolio optimization by connecting to multiple EAMs to reallocate corresponding assets. 
Ma et al. \cite{Ma2023} introduce a new MARL for optimizing financial portfolio management. The algorithm employs two agents to study the best trading policy for two distinct categories of stock trends, with a trend consistency factor that takes into account the consistency of stocks within a portfolio. Besides, the reward function now includes a novel TC regularization, which is based on the trend consistency factor value. The algorithm dynamically alternates between the two agents in order to obtain the best portfolio strategy based on the state of the market. 
Shavandi and Khedmati\cite{SHAVANDI2022118124} propose a MARL framework that leverages the collective intelligence of expert traders on various periods. The DQN and a hierarchical structure are used in the framework to train the agents. 

\textbf{Trading strategy optimization:} 
In the financial markets, developing an effective trading strategy is always a challenging issue. Traditionally, trading strategies are usually designed by individuals or teams based on their experience and skills, but there are many limitations in this approach. With the continuous advance in AI methods, MARL is widely applied in the optimization of trading strategies. It allows multiple agents to collaborate and compete to learn and improve strategies, leading to better trading results.
\cite{9931995} and \cite{qiu2021multi} worked by Qiu et al use MFRL \cite{pmlr-v80-yang18d} and MADDPG \cite{maddpg} to optimize energy trading and market strategies, respectively.
Patel \cite{patel2018optimizing} applies MARL to place limit orders to optimize market-making. The MARL framework consists of a macro-agent that decides whether to buy, sell, or hold an asset and a micro-agent that places limited orders within the order book. 
A model-free method is proposed by Karpe et al.\cite{10.1145/3383455.3422570}. It uses the Double Deep Q-Learning algorithm \cite{doubleq}, which is trained in a multi-agent realistic market simulation environment. The approach involves configuring a historical order book simulation environment with multiple agents and evaluating the simulation with real market data.
Bao \cite{bao2019fairness} proposes a MARL method to formulate stock trading strategies for investment banks with multiple clients. The method aims to balance revenue and fairness among clients with different order sizes and requirements. The proposed scheme uses RL to adapt trading strategies to complex market environments and uses MAS to optimize individual revenues. 

\textbf{Risk management:} 
Risk management is always a crucial part of business and organization management. Compared with traditional SARL, MARL can help enterprises and organizations better manage risk and reduce potential losses and risks.
Bajo et al. \cite{BAJO20126921} discuss the need for innovative tools to help small to medium enterprises predict risks and manage inefficiencies and create a multi-agent system that uses advanced reasoning to detect situations with risks and offer decision support. 
Ganesh et al. \cite{ganesh2019reinforcement} use a simulation to study how RL can be utilized for training market maker agents in a dealer market. The RL agent learns to manage inventory and adapt to market price trends while also learning about the pricing strategies of its competitors. They also propose and test reward formulations to create risk-averse RL-based market makers.
He et al. \cite{HE2023109985} propose a new approach to train a trading agent using RL by using a multi-agent virtual market model consisting of multiple generative adversarial networks. The model creates simulated market data that takes into account how the action of the agent affects the state of the market. 
A backtest of the China Shanghai Shenzhen 300 stock index futures in 2019 shows that the trained agent has a 12 percent higher profit and a low risk of loss.

\subsection{Network Security} 
Network security is an important issue facing society today, where attackers use various techniques and means to compromise computer systems and networks, threatening the security of individuals, organizations, and nations. MARL is a promising approach that can be used in the field of network security, with major applications in intrusion detection \cite{10.1007/978-3-540-87805-6_15, SETHI2021102923, 9335796, mohamed2021adversarial, louati2022distributed, louati2022distributed} and network resource optimization \cite{9348210,SUN2020107230,9254093,li2019cooperative,9329087}.
The correspondence between this application and RL methods is shown in Table \ref{network}.

\begin{table}[]
\centering
\caption{Correspondence between network security and RL methods.}
\label{network}
\begin{tabularx}{\textwidth}{|XX|c|c|c|}
\hline
\multicolumn{2}{|c|}{Applications}                                                               & Papers                                                                                     & Methods                                    & \multicolumn{1}{c|}{SA/MA} \\ \hline
\multicolumn{1}{|X|}{\multirow{6}{=}{\centering Network Security}} & \multirow{3}{=}{\centering Intrusion detection}   & \cite{10.1007/978-3-540-87805-6_15,SETHI2021102923,louati2022distributed,chowdhary2021sdn} & DQN-based \cite{dqn}                       & MA                         \\ \cline{3-5} 
\multicolumn{1}{|X|}{}                                  &                                        & \cite{9335796}                                                                             & DQN-based \cite{dqn}                       & SA                         \\ \cline{3-5} 
\multicolumn{1}{|X|}{}                                  &                                        & \cite{mohamed2021adversarial}                                                              & SARSA-based \cite{kuzmin2002connectionist} & MA                         \\ \cline{2-5} 
\multicolumn{1}{|X|}{}                                  & \multirow{3}{=}{\centering Resource optimization} & \cite{9348210,SUN2020107230,li2019cooperative}                                             & DQN-based \cite{dqn}                       & MA                         \\ \cline{3-5} 
\multicolumn{1}{|X|}{}                                  &                                        & \cite{9254093}                                                                             & MADDPG-based \cite{maddpg}                 & MA                         \\ \cline{3-5} 
\multicolumn{1}{|X|}{}                                  &                                        & \cite{9329087}                                                                             & Double-Q, AC \cite{doubleq,ac}             & MA                         \\ \hline
\end{tabularx}
\end{table}

\textbf{Intrusion detection: } 
Intrusion detection is one of the critical aspects to protect network security \cite{9705079}. However, traditional intrusion detection systems may have limitations in the face of complex and variable network attacks. MARL is an effective solution that can be used to enhance the accuracy and robustness of intrusion detection through collaborative learning and mutual communication.
Servin and Kudenko \cite{10.1007/978-3-540-87805-6_15} present a MARL-based intrusion detection method that enables the identification and prediction of normal and abnormal states in a network through learning and interaction between distributed sensors and decision-making intelligence.
Sethi et al. \cite{SETHI2021102923} propose an intrusion detection system according to MARL with attention mechanisms for efficient detection and classification of advanced network attacks. 
A DRL algorithm is proposed by Hsu and Matsuoka \cite{9335796} for the anomaly network intrusion detection systems, which can update itself to detect new types of network traffic behavior. The system is tested on two benchmark datasets and a real campus network log and compared to three classic machine learning methods and two related published results. The model is capable of processing large amounts of network traffic in real time.
Safa and Ridha \cite{mohamed2021adversarial} propose a new adversarial MARL approach-based Deep SARSA \cite{kuzmin2002connectionist} for intrusion detection in dynamic environments. The proposed algorithm addresses the problem of imbalanced distribution datasets by improving the detection of minority classes, which can improve classifier performance.
Louati et al. \cite{louati2022distributed} propose an intelligent and distributed intrusion detection system using the MAS based on parallel ML algorithms.
Chowdhary et al. \cite{chowdhary2021sdn} propose a MARL framework for an adversarial game in a software-defined network-managed cloud environment. This model takes into account the dynamic nature of the network and minimal impact on service availability.

\textbf{Resource optimization:}
Suzuki and Harada \cite{9348210} propose a safe MARL to optimize network resources efficiently even during significant changes in network demands. This method uses DRL algorithms to learn the relationship between network demand patterns and optimal allocation in advance. Safety considerations and multi-agent techniques are developed to reduce constraint violations and improve scalability, respectively. 
Sun et al. \cite{SUN2020107230} propose a dynamic controller workload balancing scheme based on MARL to address the time-consuming or under-performing issues of iterative optimization algorithms. 
Peng and Shen\cite{9254093} explore multi-dimensional resource management for UAVs in vehicular networks, and the problem is formulated as a distributive optimization problem that can be addressed by the MADDPG method \cite{maddpg}. 
Li et al. \cite{li2019cooperative} propose a MARL approach to address resource-balancing challenges within complex transportation networks. The traditional solutions leveraging combinatorial optimization face challenges due to high complexity, uncertainty, and non-convex business constraints. The proposed approach introduces a cooperative mechanism for state and reward design, resulting in better transportation which is more efficient and effective. 
Naderializadeh et al.\cite{9329087} propose a distributed resource management and interference mitigation mechanism for wireless networks using MARL. In the network, each transmitter is equipped with a DRL agent responsible for selecting the user to serve and determining the transmission power to utilize based on delayed observations from its associated users and neighboring agents.

MARL has excellent potential in the field of network security, especially when dealing with complex network attacks and defense strategies. However, there are some shortcomings of MARL in the network security domain. One of the main problems is insufficient training data and performance issues. The behaviors of attackers are usually covert and small in number, so obtaining reliable training data is a challenge. In addition, attackers may use adversarial samples to spoof MARL models, leading to model failure. Therefore, it is necessary to address the robustness and generalization problem of MARL in addition to improving its performance. The correspondence between this application and RL methods is shown in Table \ref{education}.

\subsection{Smart Education}

\begin{table}[]
\centering
\caption{Correspondence between smart education and science and RL methods.}
\label{education}
\begin{tabular}{|cl|c|c|c|}
\hline
\multicolumn{2}{|c|}{Applications}                     & Papers                                    & \multicolumn{1}{c|}{Methods} & SA/MA \\ \hline
\multicolumn{2}{|c|}{\multirow{2}{*}{Smart Education}} & \cite{asee_peer_40052,9970680}            & DQN-based \cite{dqn}         & SA    \\ \cline{3-5} 
\multicolumn{2}{|c|}{}                                 & \cite{8615217,fu2022reinforcement} & DQN-based \cite{dqn}         & SA    \\ \hline
\multicolumn{2}{|c|}{\multirow{3}{*}{RL for Science}}  & \cite{seo2021feedforward}                 & DDPG-based \cite{ddpg}       & SA    \\ \cline{3-5} 
\multicolumn{2}{|c|}{}                                 & \cite{Degrave2022}                        & AC-based \cite{ac}           & SA    \\ \cline{3-5} 
\multicolumn{2}{|c|}{}                                 & \cite{Bae2022}                            & PPO-based \cite{ppo}         & MA    \\ \hline
\end{tabular}
\end{table}

Smart education uses the IoT and AI to digitize learning processes and offer individualized learning experiences and support depending on the learning styles and features of specific students. Sensors can be used to capture students' learning behaviors and data. Communication enables real-time interaction between students and teachers, as well as collaborative learning among students. 
AI can be used to analyze learning behavior, offer personalized learning, and evaluate teaching. Scene reconstruction, experiment simulation, and remote teaching are made easier by virtual reality technology.
In MARL-based smart education, we summarize the existing techniques \cite{asee_peer_40052, 9970680, 8615217, fu2022reinforcement}. 
Education 4.0 intends to incorporate AI technology into each stage of student self-regulated learning to increase interest and effectiveness during the process \cite{HADERER20221328, SCHUMACHER2021100791, 8990763}. 
Tang and Hare \cite{asee_peer_40052} create an adaptive tutoring game that allows students to personalize their learning without the guidance of teachers. In order to optimize student learning, this system uses a Petri net graph structure to monitor students' progress in the game and an RL agent to adaptively change system behavior in response to student performance.
Then they apply Petri Nets and hierarchical reinforcement learning algorithm to personalized student assistance based on the above game \cite{9970680}. The algorithm can assist teachers in giving students in-game instruction and feedback that is specifically tailored to them, allowing them to gradually master complex knowledge and skills by breaking down the tasks in games into several stages. The algorithm can help educators provide customized support and feedback to students in games and gradually master complex knowledge and skills by dividing the tasks in games into multiple levels. \cite{8615217} and \cite{fu2022reinforcement} both monitor student learning progress using data gathered by sensors and offer students personalized learning advice using RL techniques. 

Smart Education based on MARL can enhance teaching efficiency, save time, and ultimately, better learning outcomes for students. However, the collection of daily behavioral data from students is required by smart education, which presents privacy concerns. Additionally, since the core of intelligent education is human, its purpose is to assist teachers in teaching and students in learning. As a result, it necessitates prioritizing switching according to different scenarios, such as when there are discrepancies between the assessment of the teacher and AI for the level of knowledge mastery. Improper prioritization switching may lead to reduced educational effectiveness and poor student experiences. Therefore, how to conduct reasonable prioritization switching is a problem that needs to be explored.

\subsection{RL for Science}
Recently, AI for science has been a popular topic, and AI is highly regarded as a critical tool in achieving scientific progress \cite{9966863}. RL has demonstrated significant scientific potential, with particular promise in chemistry, physics, and materials research. RL has proven instrumental in solving challenges like exploring uncharted physical phenomena. The correspondence between this application and RL methods is shown in Table \ref{education}.
Seo et al. \cite{seo2021feedforward} utilize RL to control feedforward $\beta$ in the KSTAR tokamak.
Degrave et al. \cite{Degrave2022} introduce an innovative RL approach that enables the magnetic control system of a tokamak fusion device to learn autonomously, achieving precise control over various plasma configurations, significantly reducing design efforts, and representing a pioneering application of RL to the fusion domain.
Bae et al. \cite{Bae2022} introduce a scientific MARL (SciMARL) for discovering wall models in turbulent flow simulations, dramatically reducing computational cost while reproducing key flow quantities and offering unprecedented capabilities for simulating turbulent flows.
RL scientific research offers more possibilities, and we believe that RL will have a wider range of scientific applications in the future.

\section{Visionary Prospects} \label{Visionary}
Although MARL has shown superior performance in many domains, some issues, such as safety, robustness, and generalization, limit the application of MARL in the real world. We believe that maximizing the superiority of MARL in practical applications in the future needs to first address these issues and need to consider the moral constraints of human society. 
This section reviews the current state of research in four areas: safety, robustness, generalization, and ethical constraints, and discusses the gaps that need to be addressed in future research.

\subsection{Safety of Multi-agent Reinforcement Learning}
The increasing popularity of MARL has brought attention to the need to ensure the safety of these systems. In MARL, the actions of one agent can potentially cause harm to the task or other agents involved. Therefore, there is a pressing need to develop safe MARL approaches. To achieve safety in MARL, one common approach is to add constraints to the training process. By incorporating safety constraints, agents are encouraged to avoid unsafe actions that could lead to task failure or harm to other agents. 
There have been numerous reviews on the safety of RL, as summarized in \cite{JMLR:v16:garcia15a}, \cite{gu2022review}, and \cite{trustRL}. However, there is currently no systematic review of the safety of MARL, and there is relatively little research on this topic. In this section, we give a definition of safe MARL which is used in \cite{gu2021multi}.

\begin{figure}
    \centering
    \includegraphics{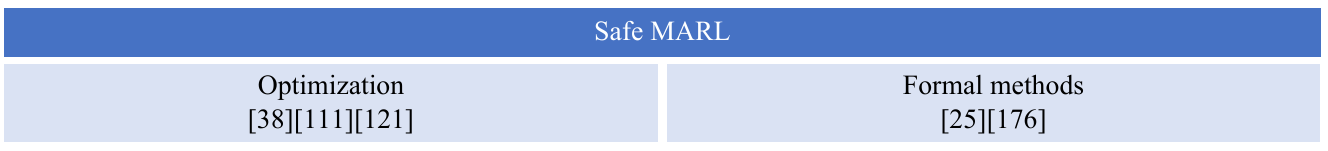}
    \caption{Categories of Safety in MARL}
    \label{safe}
\end{figure}

\begin{definition}[Safe MARL]
    A multi-agent constrained stochastic game can be modeled as the tuple 
    $$
        \left<N, \mathcal S, \mathcal A^1, \cdots, \mathcal{A}^N, R, \mathcal C^1, \cdots, \mathcal C^N, \bm c^1,\cdots, \bm c^N, p, \gamma\right>,
    $$
    where $R: \mathcal S \times \mathcal A^1 \times \cdots \times \mathcal A^N \times \mathcal S \rightarrow \mathbb R $ is the joint reward function, $\mathcal C^i = \{C^i_j\}^{i \le N}_{1\le j \le m^i}$ is a set of cost function of agent $i$ ($m^i$ is the number of cost functions of agent $i$), $C^i_j: \mathcal S \times \mathcal A^1\times \cdots \times \mathcal A^N \times \mathcal S \rightarrow \mathbb R $ is the cost function, and $\bm c^i = \{c^i_j\}^{i \le N}_{1\le j \le m^i} \in \mathbb R$ is cost-constraining values.
\end{definition}
The goal of agents is to maximize the expected total reward while trying to satisfy the safety constraint of each agent,
\begin{equation}   \label{safety}
\begin{aligned}
    &\mathcal J\left(\bm \pi \right) = \mathbb E_{\bm \pi}\left[\sum_{t=0}^\infty \gamma^tR\left(s_t,\bm a_t,s_{t+1}|s_0 = s\right)\right], \\
    s.t. &\mathcal J^i_j\left(\bm \pi \right) = \mathbb E_{\bm \pi}\left[\sum_{t=0}^\infty \gamma^tC^i_j\left(s_t,\bm a_t,s_{t+1}|s_0 = s\right)\right] \le c^i_j, \\
    & \qquad \qquad  \forall j = 1,\cdots, m^i.
\end{aligned} 
\end{equation}
We then summarize relevant research from two perspectives: optimization and formal methods, as shown in Fig.\ref{safe}. 
\subsubsection{Optimization}
Gu et al. \cite{gu2021multi} introduce Multi-Agent Constrained Policy Optimization (MACPO) and MAPPO-Lagrangian to devise safety MARL algorithms. These algorithms aim to meet safety constraints while concurrently enhancing rewards by integrating theories from constrained policy optimization and multi-agent trust region learning, yielding strong theoretical guarantees. Furthermore, the authors have established a benchmark suite, Safe Multi-Agent MuJoCo, to evaluate the efficacy of their approaches, which exhibit performance levels comparable to baselines and persistently comply with safety constraints.
Lu et al. \cite{lu2021decentralized} propose a method called Safe Decentralized Policy Gradient (Safe Dec-PG) to solve a distributed RL problem where agents work together with safety constraints. The method is decentralized and considers coupled safety constraints while ensuring a measurable convergence rate. It can also solve other decentralized optimization problems.
Liu et al. \cite{10.1007/978-3-030-86486-6_10} propose a novel algorithm CMIX that can be used for MARL in a partially observable environment with constraints on both peak and average reward. CMIX enables CTDE and outperforms existing algorithms in maximizing the global reward function subject to constraints. The algorithm is evaluated on two scenarios, including a blocker game and a vehicular network routing problem, demonstrating its ability to satisfy both peak and average constraints, which has not been achieved before in a CTDE learning algorithm. 
\subsubsection{Formal methods}
MARL is being used in safety-critical applications, but current methods do not guarantee safety during learning. To address this, two approaches for safe MARL have been presented in \cite{elsayed2021safe}: centralized shielding monitors actions of all agents and corrects unsafe actions, while factored shielding uses multiple shields to monitor subsets of agents concurrently. Both approaches ensure safety without sacrificing policy quality, but factored shielding is larger numbers of agents.
Sheebaelhamd et al. \cite{sheebaelhamd2021safe} improve the MADDPG framework for multi-agent control problems with safety constraints. A safety mechanism is integrated into the deep policy network to avoid in-feasibility problems in the action correction step, which guarantee constraint satisfaction using exact penalty functions. Empirical results show that this approach reduces constraint violations, enabling safety during learning.
\subsubsection{Limitations of current methods}
Although there has been some progress in researching the safety of MARL, there are still some limitations. First, the existing approach to MARL safety is designed for small numbers of agents and may not be applicable to large-scale systems. Second, most existing research on MARL safety assumes that the environment is static and unchanging. In real-world applications, however, the environment is often dynamic and unpredictable, which can pose additional safety risks. Finally, in order to apply MARL to human society, it is necessary to add constraints to protect human safety. Furthermore, human interactions lead to a non-Markov environment. Hence, MARL which accounts for the safety of large-scale human society, is a challenging and significant research direction for the future.

\subsection{Robustness of Multi-agent Reinforcement Learning}

The robustness of DL in classification tasks has a series of studies \cite{goodfellow2014explaining, 7467366, huang2017adversarial, jiao2022asynchronous, jiao2022distributed}. RL is a sequential decision problem, where misclassification at a one-time step is not equivalent to expecting the minimum reward. In MARL, a decision failure of any agent can lead to team task failure, which makes the study of robustness MARL challenging. Furthermore, MARL faces various challenges in real-world applications, such as uncertainty in the environment, uncertainty policies of other agents, and sensor noise. All these factors may cause the trained models to perform poorly or fail. Therefore, it is crucial to improve the robustness of MARL, which will help ensure that the models can operate stably and reliably in various situations. The following are related definitions of robust MARL. We use the definition of \cite{zhou2022romfac} and \cite{,NEURIPS2020_77441296}.
\begin{definition}[Robustness against state observations perturbation]
    A state-adversarial stochastic game can be defined as a tuple $$\left<\mathcal S, \mathcal A^1,\dots,\mathcal A^N,\mathcal B^1,\dots, \mathcal B^M, R^1,\dots, R^N,p,\gamma \right>$$ where $\mathcal B^j$ is the uncertainty set of adversarial states of agent $j$, and $M$ is the number of attacked agents such that $M \le N$.
\end{definition}
Given the joint policy $\bm \pi: \mathcal S \rightarrow \mbox{PD} \left(\mathcal A^1\times \cdots \times \mathcal A^N \right)$ and the joint adversarial perturbation $\bm v:\mathcal S \rightarrow \mathcal B^1\times \cdots \times \mathcal B^M $,  The Bellman equation with fixed $\bm \pi$ and $\bm v$ is as follows,
	\begin{equation} \label{bv}
		\begin{aligned}
			\hat{V}^i_*(s)=\mathop {max} \limits_{\pi^i\left(\cdot|s\right)} \mathop{min} \limits_{v}
   \sum_{\bm a  \in \mathcal A^1  \times \cdots  \times  \mathcal A^N} \bm \pi\left(\bm a|s ,\bm v(s)\right)\sum\limits_{s' \in \mathcal S}\left( p\left(s'|s,\bm a\right) 
\left(R^i\left(s,\bm a,s'\right)+\gamma\hat{V}^i_*(s')\right)\right),
		\end{aligned}
	\end{equation}%
\begin{definition}[Robustness against model perturbation]
    A model-adversarial stochastic game can be defined as the tuple
    $$
        \left<N, \mathcal{S}, \mathcal A^1, \cdots, \mathcal A^N, \hat{\mathcal {R}}^1, \cdots, \hat{\mathcal {R}}^N, \hat{\bm p},\gamma \right>,
    $$
    where $\hat{\mathcal {R}}^i$ and $\hat{\bm p}$ are the uncertainty sets of reward functions and transition probabilities, respectively. 
\end{definition}
The Bellman-type equation is as follows:
\begin{equation}
    \hat V^i_*\left(s\right) = \mathop {max} \limits_{\pi^i\left(\cdot|s\right)} \mathop {min} \limits_{\hat R^i \in \hat{\mathcal {R}}^i, \hat p \in \hat{\bm p}} \sum_{\bm a\in \mathcal A^1\times\cdots\times\mathcal A^N}\bm \pi\left(\bm a|s\right)\sum_{s'\in \mathcal S}\left(\hat p \left(s'|s,\bm a\right)  \hat R^i\left(s, \bm a, s'\right) + \gamma \hat V^i_*\left(s'\right) \right)
\end{equation}

Currently, research on the robustness of MARL is being pursued from both attacks and defense. Attacks research aims to identify stronger perturbations to test the robustness of MARL models, while defense aims to develop MARL algorithms that are robust to perturbations. 
\begin{figure}
    \centering
    \includegraphics{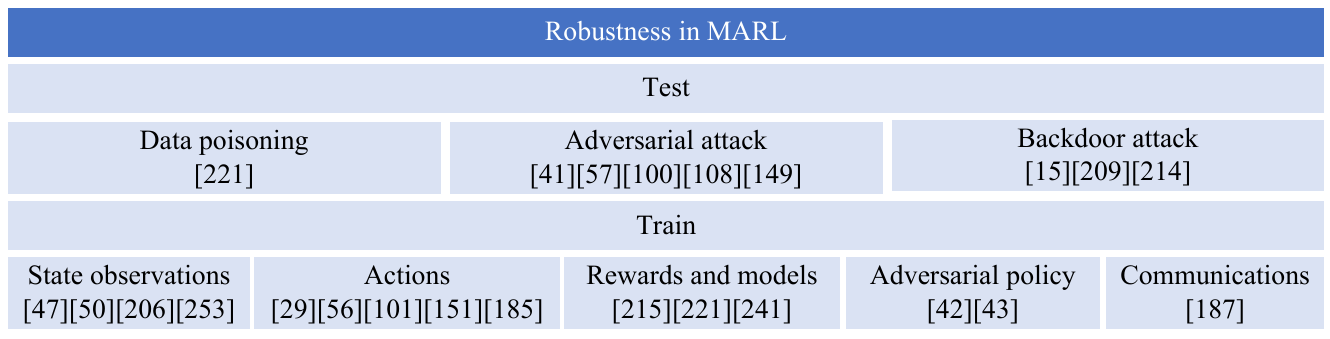}
    \caption{Categories of Robustness in MARL}
    \label{Robustness}
\end{figure}

\subsubsection{Testing}: 
As shown in Fig. \ref{Robustness}, similar to DL, the robustness testing methods for MARL can be classified into three categories: adversarial attacks, backdoor attacks, and data poisoning.

\textbf{Data poisoning:} Wu et al. \cite{wu2022reward} discuss how an attacker can modify rewards in a dataset used for offline MARL to encourage each agent to adopt a harmful target policy with minimal modifications. 
The attacker can establish the target policy as a Markov perfect dominant strategy equilibrium, which is a strategy that rational agents will adopt.
The article explores the effectiveness of attacks on various MARL agents and their cost compared to separate single-agent attacks. It also examines the relationship between dataset structure and attack cost and highlights the need for future research on defense in offline MARL.

\textbf{Adversarial attacks:} Lin et al. \cite{9283830} show that Cooperative MARL (c-MARL) is vulnerable to attacks on a single agent. By manipulating agent observations, the attacker reduces the overall team reward. 
The proposed attack strategy involves training a policy network to induce the victim agent to take an incorrect action and utilizing targeted adversarial attack methods to compel the agent to take that action.
Experiments demonstrate a significant reduction in team reward and winning rate.
Guo et al. \cite{Guo_2022_CVPR} discuss the potential vulnerabilities of c-MARL algorithms to adversarial attacks and the importance of testing their robustness before deployment in safety-critical applications. The authors propose MARLSafe, a comprehensive testing framework that considers state, action, and reward robustness to address this. Experimental results on the SMAC environment demonstrate the low robustness of advanced c-MARL algorithms in all aspects.
Hu and Zhang \cite{hu2022sparse} propose a sparse adversarial attack on c-MARL systems to test their robustness. The attack is trained using MARL with regularization and is shown to significantly decrease performance when only a few agents are attacked at a few timesteps. This highlights the need for more robust cMARL algorithms.
Pham et al. \cite{pham2022evaluating} introduce a novel model-based approach for evaluating the robustness of c-MARL agents against adversarial states. They demonstrate the superiority of their approach over existing baselines by crafting more robust adversarial state perturbations and employing a victim-agent selection strategy. 
Through experiments on multi-agent MuJoCo benchmarks, they demonstrate that 
the approach is effective by achieving a reduction in total team rewards.
Li et al. \cite{li2023attacking} propose the Adversarial Minority Influence attack, which introduces an adversarial agent that influences other cooperative victims to achieve worst-case cooperation. The attack addresses the complexity and cooperative nature of c-MARL by characterizing and maximizing the influence from the adversary to the victims. The proposed approach is demonstrated to be superior to existing methods in various simulation environments.

\textbf{Backdoor attack: } Chen et al. \cite{chen2022marnet} introduce a novel backdoor attack framework, known as MARNet, which is specifically designed for c-MARL scenarios. MARNet comprises three primary modules: trigger design, action poisoning, and reward hacking, all of which work together to manipulate the actions and rewards of poisoned agents. The framework is evaluated on two popular c-MARL algorithms, VDN \cite{vdn} and QMIX \cite{qmix}, in two commonly used c-MARL games. The experimental results demonstrate that MARNet outperforms baselines from SARL backdoor attacks, reducing the utility under attack by up to 100\%. Although fine-tuning is employed as a defense mechanism against MARNet, it is not entirely effective in eliminating the impact of the attack.
Wang et al. \cite{9541185} investigate research on the backdoor attack for DRL-based Autonomous Vehicles (AVs) controllers. They develop a trigger based on traffic physics principles. Experiments are conducted on both single-lane and two-lane circuits, and they demonstrate that the attack can cause a crash or congestion when triggered while maintaining normal operating performance. These findings underscore the importance of robust security measures in AVs controller design.
Wang et al. \cite{wang2021backdoorl} examine backdoor attacks in MARL systems and put forward a technique called BACKDOORL to detect and prevent such attacks.

\subsubsection{Training}: Robustness testing and training in MARL are still in the early stages of research. Therefore, we summarize robustness training methods from five aspects: state observation, action, reward and model, adversarial policy, and communication.

\textbf{State Observations:} In our previous work\cite{zhou2022romfac}, 
we combine a policy gradient function and an action loss function, along with a regularized action loss term, to develop a new objective function for training actors in mean-field actor-critic reinforcement learning \cite{pmlr-v80-yang18d} that improves its robustness. Furthermore, we define State-Adversarial Stochastic Game (SASG) and discuss its properties. 
Due to the traditional solution concepts do not always exist in SASG, \cite{han2022solution} and \cite{he2023robust} introduce a new solution concept called robust agent policy and develop a Robust Multi-Agent Adversarial Actor-Critic (RMA3C) algorithm to learn robust policies for MARL agents.
Wang et al. \cite{wang2023robust} propose a training framework for c-MARL to address the weakness of agents to adversarial attacks. The framework generates adversarial attacks on agent observations to help them learn a robust cooperative policy. The attacker selects an agent to attack and outputs a perturbation vector. The victim policy is then trained against the attacker. Experimental results demonstrate that the generated attacks improve the robustness against observation perturbations. 

\textbf{Actions:} Foerster et al. \cite{foerster2017learning} consider how the policies adopted by different agents in the environment interact with each other and affect the learning process of all agents. They propose Learning with Opponent-Learning Awareness (LOLA), a framework that takes into account the influence of one agent's policy on the expected parameter update of the other agents through a specific term.
The method leads to the emergence of cooperation in the iterated dilemma of prisoners and convergence to the Nash equilibrium in repeated matching pennies. 
The extension of the policy gradient estimator enables efficient computation of LOLA, making it suitable for handling large parameters and input spaces that use nonlinear function approximators. 
Li et al. \cite{Li_Wu_Cui_Dong_Fang_Russell_2019} design MiniMax Multi-agent Deep Deterministic Policy Gradient (M3DDPG) to train MARL agents with continuous actions that can handle robustness issues in complex environments. M3DDPG  adds a minimax component to MADDPG \cite{maddpg} and employs multi-agent adversarial learning (MAAL) to optimize the learning objective. 
Through experiment evaluation in four multi-agent environments, the proposed algorithm surpasses existing baselines in terms of performance.
Based on \cite{Li_Wu_Cui_Dong_Fang_Russell_2019}, Sun et al. \cite{9812321} apply the convex relaxation of neural networks instead of MAAL to apply the convex relaxation of neural networks to overcome computationally difficult,  which enables robustness in interacting with agents that have significantly different behaviors and achieves a certified bound of the original optimization problem.
To overcome the computational difficulties of MAAL, Sun et al. \cite{9812321} utilize the convex relaxation technique to guarantee robustness in the interaction of agents with varying actions and yield a certified bound for the original optimization problem.
Phan et al. \cite{Phan_Belzner_Gabor_Sedlmeier_Ritz_Linnhoff-Popien_2021} propose a value decomposition scheme that trains competing teams of varying sizes to improve resilience against arbitrary agent changes. By doing so, RADAR offers a more versatile and flexible approach to MARL that can adapt to changing agent behavior and system conditions. 
According to \cite{hu2021robust}, in order to enhance robustness, non-adversarial agents should collaborate and make decisions based on correlated equilibrium rather than acting independently. The authors introduce new approaches to encourage agents to learn and follow correlated equilibrium while maintaining the benefits of decentralized execution.

\textbf{Rewards and models:} Wang and Zou \cite{NEURIPS2021_3a449677} propose a sample-based approach to estimate the uncertainty set of a misspecified MDP in model-free robust RL. They develop robust Q-learning and robust TDC algorithms that converge to optimal or stationary points without additional conditions on the discount factor. The algorithms also have similar convergence rates as their vanilla counterparts and can be extended to other RL algorithms.
Zhang et al. \cite{NEURIPS2020_77441296} focus on the problem of MARL in situations where there is uncertainty in the model, such as inaccurate knowledge of reward functions. They model this as a robust Markov game, where agents aim to find policies that lead to equilibrium points that are robust to model uncertainty. 
They present a novel solution concept known as robust Nash equilibrium and a Q-learning algorithm that guarantees convergence. Additionally, policy gradients are derived, and an actor-critic algorithm that uses function approximation is developed to effectively tackle large state-action spaces.
Wu et al. \cite{wu2022reward} introduce linear programs that can efficiently address the attack problem and analyze the connection between the characteristics of datasets and the minimal attack cost.

\textbf{Adversarial policy:} Guo et al. \cite{guo2022backdoor} propose Backdoor Detection in MARL systems, using Policy Cleanse to detect and mitigate Trojan agents and their trigger actions. Besides, they also design a machine unlearning-based approach to effectively mitigate the detected backdoors.
In contrast to previous techniques that rely on the zero-sum assumption, the recent work by Guo et al. \cite{pmlr-v139-guo21b} proposes a novel approach that resets the optimization objective and employs a new surrogate optimization function. This method has been shown through experiments to significantly enhance the ability of adversarial agents to exploit weaknesses in a given game and take advantage of any inherent unfairness in the game mechanics. Moreover, agents that are trained adversarially against this approach have demonstrated a greater level of resistance against adversarial attacks. Overall, these findings suggest that the proposed approach represents a promising direction for improving the robustness and fairness of game-playing AI agents.

\textbf{Communication: } A certifiable defense mechanism is proposed by Sun et al. \cite{sun2023certifiably}, which employs a message-ensemble policy to merge several message sets with random ablations. Theoretical analysis indicates that this mechanism can withstand various adversarial communications.

\subsubsection{Limitations of current methods}
Current research on MARL robustness leaves much to be desired. First, the recent research only focuses on one of the states, actions, or policies and needs to consider the robustness of a combination of multiple aspects. Second, there needs to be more team robustness evaluation metrics. It is insufficient to test the robustness of MARL only by way of attacks because it can only cover some possible perturbations. In addition, existing studies tend to ignore the impact of non-cooperative and malicious behaviors caused by human factors on robustness, which is also an issue that needs further research. Therefore, further in-depth integration of robust MARL with multiple perturbation types, verifiable robustness evaluation metrics, and robustness with human intervention must be considered in the future.

\subsection{Generalization of Multi-agent Reinforcement Learning}
\begin{figure}
    \centering
    \includegraphics{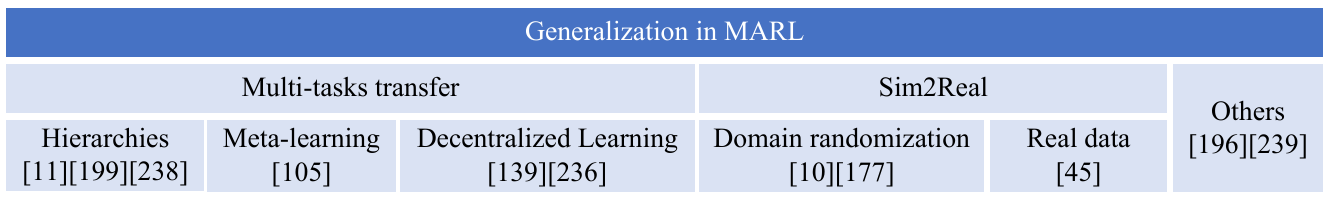}
    \caption{Categories of Generalization in MARL}
    \label{generalization}
\end{figure}

Within the domain of MARL, generalization pertains to the capacity of agents to transfer their learned knowledge and skills from a specific environment or scenario to novel and diverse ones without necessitating significant modifications or retraining. Several surveys have investigated generalization in RL \cite{trustRL, 9308468, electronics9091363, 10.1613/jair.1.14174}. In the generalization of SARL, various techniques such as domain randomization \cite{7354126,rajeswaran2016epopt,Sadeghi_2018_CVPR}, causal inference \cite{ke2019learning, scherrer2021learning, pmlr-v119-zhang20t}, and meta-learning \cite{pmlr-v70-finn17a, 9196540,pmlr-v80-kaplanis18a} have been employed to address generalization issues. However, compared to single-agent settings, research on the generalization of MARL remains relatively scarce. In this regard, we provide an overview of pertinent work from two perspectives, namely multi-task learning, and sim2real, as shown in Fig. \ref{generalization}. 


\subsubsection{Multi-tasks transfer}
The goal of multi-task learning is to improve the generalization ability of a model by incorporating knowledge from related tasks as a form of inductive bias. In order to learn shared and task-specific representations and improve overall performance and efficiency in complicated domains entails training a model to carry out several tasks at once.

\textbf{Hierarchies: } To address the issue of generalization to unknown opponents in multi-agent games, Vezhnevets et al. \cite{pmlr-v119-vezhnevets20a} propose a hierarchical agent architecture grounded in game theory, which enables credit assignment across hierarchy levels and achieves better generalization to unseen opponents than conventional baselines.
Carion et al. \cite{NEURIPS2019_3c3c139b} propose a structured prediction method to assign agents to tasks that uses coordination inference procedures and scoring models. 
Zhang et al. \cite{zhang2023discovering} propose an offline multi-task collaborative reinforcement learning algorithm called ODIS, which is able to extract universal coordination skills from offline multi-task data, enabling better generalization in handling multi-task coordination problems. 
Specifically, the ODIS algorithm has a two-step process for improving the generalization and performance of c-MARL tasks.
First, it extracts coordination skills from offline data that are applicable across different tasks. It then uses these skills to differentiate between different agent behaviors.
Second, it trains a coordination policy that selects the most effective coordination skills using the CTDE paradigm. The effectiveness of ODIS is demonstrated in experiments where it significantly improves generalization to unseen tasks, achieving superior performance in various cooperative MARL benchmarks. Importantly, the ODIS algorithm achieves these results using only limited sources of offline data.

\textbf{Meta-learning: }
Liang et al. \cite{Liang2022-yf} present a Self-adaptive Meta-learning (SAML) framework that employs gradient-based methods to combine individual task policies into a unified policy capable of adapting to new tasks.
Experimental results demonstrate that SAML outperforms baseline methods in terms of efficiency and continuous adaptation. 

\textbf{Decentralized learning: } Omidshafiei et al. \cite{pmlr-v70-omidshafiei17a} tackle the challenge of multi-task MARL with partial observability and limited communication. They introduce a decentralized single-task learning approach that can be synthesized into a unified policy for multiple correlated tasks without the need for explicit indication of task identity.
The work by Zeng et al. \cite{pmlr-v161-zeng21a} presents a novel mathematical framework for addressing multi-task RL problems using a policy gradient method. Specifically, the authors propose a decentralized entropy-regularized policy gradient method for solving these problems. The efficacy of the proposed method is evaluated through experimental results on both small-scale and large-scale multi-task RL problems. The findings demonstrate that the proposed approach offers promising performance for tackling complex multi-task RL problems.

\subsubsection{Sim2Real} 
To train MARL agents, simulations are often used due to their efficiency and ease of implementation. However, a significant challenge arises when attempting to transfer policies learned in simulation to the real world, as differences between the two environments can lead to a performance gap.
To address this issue, researchers have been investigating methods for Sim2Real transfer, which aim to minimize the performance gap between simulation and the real world. These methods typically involve fine-tuning policies in the real world, using domain randomization to increase the generalization of policies learned in simulation, or combining real data to achieve better results.

\textbf{Domain randomization: }Candela et al. \cite{9981319} create a simulation platform for autonomous driving and use the MAPPO with domain randomization to enable the transfer of policies from simulation to reality. In our previous work \cite{9993797}, we developed a simulation platform for multi-UAV transport, utilizing domain randomization to facilitate the transfer from simulation to reality. Additionally, we formulated a non-stationary variant of Markov games and established the efficacy of RNNs in addressing non-stationary Markov games.

\textbf{Real data: }Gurevich et al. \cite{gurevichreal} 
present a novel approach for implementing homogeneous MAS by transferring data between real and simulated robots. Their method involves designing a deep neural network architecture called CR-Net, which can simulate the motion of individual robots in this system. To train the CR-Net in a simulated environment, they generate synthetic data using a generative model trained on real data from one robot. The effectiveness of their approach is validated by testing the RL models trained using this method on real ground and underwater vehicles, which showed successful policy transfer from simulation to reality. 

\subsubsection{Others}
The generalization to unexplored state-action pairs is considered in \cite{van2021model}, which uses tensors of low CP-rank to model the transition and reward functions.
Zhang et al. \cite{pmlr-v157-zhang21b} propose a novel multi-task actor-critic paradigm based on a share critic with knowledge transfer to solve heterogeneous state-action learning problems.

\subsubsection{Limitations of current methods}
Current research in multi-agent learning has mainly focused on generalization in the context of cyber-physical systems, which considers the abstraction of agents to unknown agents and the differences between virtual and real-world environments. However, the functionality of human social systems is multifaceted, and human behavior is highly diverse, making the consideration of the generalization of interactions with humans a crucial research area for MAS. For instance, in intelligent transportation systems, traffic signal control algorithms based on MARL must generalize over different cities and time periods. Similarly, in smart education, personalized education assistance based on MARL needs to consider individual differences in living environments and personality traits to develop tailored learning plans for students. Hence, MARL which accounts for the generalization of human behavior, is a promising and challenging research direction for the future.
\subsection{Learning with Ethical Constraint}
\begin{figure}
    \centering
    \includegraphics{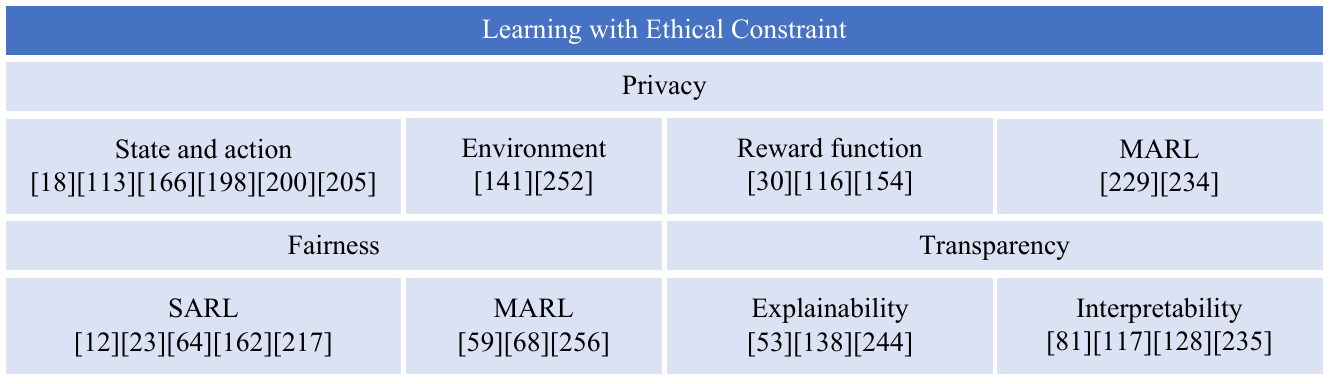}
    \caption{Categories of MARL with Ethical Constraint}
    \label{ethical}
\end{figure}

As AI technology continues to evolve, it is increasingly important to consider the ethical implications of AI systems \cite{ASHOK2022102433}. MARL systems involve the interaction of multiple agents whose actions can have significant real-world. Therefore, it is critical to ensure that the design and training of MARL systems take ethical considerations into account. We summarize research related to the ethical constraints of MARL in terms of privacy protection, fairness, and transparency, as shown in Fig.\ref{ethical}.
\subsubsection{Privacy protection}

Privacy protection is a long-standing issue extensively discussed in machine learning. Some of the main topics and techniques studied in this area include differential privacy, federated learning, cryptography, trusted execution environments, and ML-specific approaches\cite{de2020overview}. The research on privacy protection in RL is still in its early stages. We outline relevant studies in the following areas: the privacy of state and action, environment, reward function, and MARL scenario. 



\textbf{State and action: }Venkitasubramaniam \cite{6736549} proposes an MDP to explore the development of controller actions while satisfying privacy requirements. They analyze the balance between the achievable privacy level and system utility using analytical methods. The optimization problem is formulated as a Bellman equation which owns the convex reward functions for a certain category of MDPs, and as a POMDP with belief-dependent rewards for the general MDP with privacy constraints.
Differentially private algorithms are used in protecting reward information by Wang et al .\cite{NEURIPS2019_6646b06b} for RL in continuous spaces. The authors propose a method for protecting the value function approximator, which is realized by incorporating functional noise iterative into the training. They provide rigorous privacy guarantees and gain insight into the approximate optimality of the algorithm. Experiments show improvement over existing approaches.
Vietri et al. \cite{pmlr-v119-vietri20a} use the notion of Joint Differential Privacy (JDP) and a private optimism-based learning method to address the privacy problem for episodic RL. Chowdhury et al. \cite{Chowdhury_Zhou_2022} 
design two frameworks, i.e., policy optimization and value iteration, and not only consider the JDP but Local Differential Privacy (LDP) in finite horizon tabular MDP to minimize regret. The previous text describes the use of differential privacy as a means of protecting sensitive user data in RL. There are also other methods of protecting user privacy, such as cryptographic techniques. Sakuma et al. \cite{10.1145/1390156.1390265} use a homomorphic encryption algorithm to realize the privacy protection of distributed RL. They divide private information based on time and observation, design a sarsa privacy protection method based on random actions for these two division methods, and extend these to Q-learning based on greedy and $\epsilon$-greedy action selections. 
A new privacy-preserving RL method is proposed by Liu et al. \cite{8630059}
named Preyer to provide treatment options for patients while protecting their privacy. Preyer is composed of an innovative encrypted data format, a secure mechanism for plaintext length management, and a privacy-preserving RL with experience replay.

\textbf{Environment: } Pan et al. \cite{10.5555/3306127.3331715} first investigate the privacy in RL environment. They propose two methods based on genetic algorithms and shadow policies, respectively. Zhou \cite{10.1145/3508028} first discusses how to achieve privacy protection in finite-horizon MDPs, which have large state and action spaces. The author proposes two privacy-preserving RL algorithms according to value iteration and policy optimization and proves that they can achieve sub-linear regret performance while ensuring privacy protection. 

\textbf{Reward: } Fu et al. \cite{fu2017learning} and Prakash et al. \cite{Prakash_Husain_Paruchuri_Gujar_2022} investigate the problem of how to preserve the privacy of reward functions in reinforcement learning by employing adversarial reward and inverse reinforcement learning techniques.
Liu et al. \cite{liu2021deceptive} studies privacy-preserving RL using dissimulation to hide the true reward function. Two models are presented and evaluated through computational and human experiments, showing that resulting policies are deceptive and make it more difficult for observers to determine the true reward function. 

\textbf{MARL: } The differential privacy is used by Ye et al. \cite{9170873} in the field of multi-agent planning for the first time to protect agent privacy. Based on differential privacy, they propose a new strong privacy-preserving planning method, which can not only ensure strong privacy but also control communication overhead. 
Yuan et al. \cite{yuan2022pp} delve into the issue of integrating Cooperative Intelligence (CI) to enhance the efficiency of communication networks, which is hampered by privacy concerns and practical limitations in communication. In response, the authors present a Privacy-preserving scheme based on MARL (PP-MARL) that employs a HE-friendly architecture.  Experiment results indicate that PP-MARL exhibits better performance in privacy protection and reduced overhead compared to state-of-the-art approaches. Nonetheless, preserving privacy in CI-enabled communication networks remains a formidable challenge, especially when the number of agents involved is subject to variation or the system scales up. The research on privacy protection for MARL is still limited, and some studies have explored the use of differential privacy techniques to enhance the performance of MARL \cite{9269516,cheng2022multi} or against malicious advise \cite{8685696}.

\subsubsection{Fairness}

Fairness in machine learning refers to the concern that machine learning models and algorithms should not discriminate or create bias against certain groups of people based on their protected characteristics, such as race, gender, age, religion, etc. The review paper \cite{10.1145/3494672} provides a comprehensive summary of existing techniques. However, research on fairness in RL is still limited, and we provide an overview from both single-agent and multi-agent perspectives. 

\textbf{SARL: }  Jabbari et al.\cite{pmlr-v70-jabbari17a} first consider the fairness in RL and demonstrate that an algorithm conforming to their fairness constraint requires an exponential amount of time to achieve a non-trivial approximation to the optimal policy. To overcome this challenge, they introduce a polynomial time algorithm that satisfies an approximate form of the fairness constraint.
Weng et al. \cite{weng2019fairness}  address the issue of complete unfairness for some users or stakeholders by using a social welfare function encoded with fairness. 
Chen et al. \cite{9488823} introduce a novel approach to incorporate fairness in actor-critic RL for network optimization problems. By considering the shape of the fairness utility function and past reward statistics, their proposed algorithm adjusts the rewards using a weight factor that is dependent on both of these factors.
Ren et al. \cite{xx2022108242} propose a novel framework to obtain optimum and relative fairness solutions in space applications, including a new image quality representation method, a finite MDP model, and an algorithm based on RL. 
Deng et al. \cite{deng2022reinforcement} propose an RL algorithm that enforces stepwise fairness constraints to ensure group fairness at every time step.

%
\textbf{MARL: }  Jiang and Lu \cite{NEURIPS2019_10493aa8} propose a hierarchical RL model, named FEN, which is aimed at both obtaining fairness and efficiency objectives. FEN decomposes fairness for each agent and utilizes a structure with a high-level controller and multiple sub-policies to avoid multi-objective conflict. The study by Zimmer et al. \cite{pmlr-v139-zimmer21a} also focuses on the two aspects of fairness and efficiency. They propose a generic neural network architecture to address this problem, which consists of two sub-networks specifically designed to consider the two aspects of fairness and can be implemented in centralized training and decentralized execution or fully decentralized MARL settings. 
In multi-intersection scenarios, Huang et al. \cite{huang2023fairnessaware} propose a novel fairness-aware model-based MARL (FM2Light) to deal with unfair control with superior reward design.

 
\subsubsection{Transparency}
Transparency is essential for building reliable MARL decision systems. Decision-making interactions among multiple agents are very complex and difficult to understand and explain. Without a transparent understanding of the interactions and decision-making processes among agents, the reliability and trustworthiness of the system are affected. Therefore, studying the transparency of MARL is an important direction. We summarize it in terms of both explainability and interpretability.

\textbf{Explainability} refers to the ability of a model in machine learning to provide a rationale for its outputs that can be easily comprehended and trusted by humans \cite{10.1007/978-3-030-57321-8_5,10.1145/3527448}.
Heuillet et al. \cite{9679742} use a game theory concept of shapley values to explain the contribution of one agent in MARL and use Monte Carlo sampling to approximate shapley values to overcome the high overhead. This method provides an explanation for the model but can not give the precise reason why the action is taken by the agent.
Ohana et al. \cite{10.1007/978-3-030-82017-6_12} also use shapley values
to understand the model behavior and explain local feature contributions.
Zhang et al. \cite{ZHANG2021383} propose a framework composed of a variational autoencoder and graph neural networks to encode the interactions between pairs of agents.

\textbf{Interpretability} refers to the ability of a human to understand and explain the inner workings of a machine learning model \cite{10.1007/978-3-030-57321-8_5,10.1145/3527448}.
Kazhdan et al. \cite{9207564} develop a library named MARLeME which uses symbolic models to improve the interpretability of MARL. It can be employed across a broad spectrum of existing MARL systems and has potential applications in safety-critical domains.
Liu et al. \cite{liu2022mixrts} propose a novel interpretable architecture based on soft decision trees with recurrent structure. 
Milani et al. \cite{10.1007/978-3-031-26412-2_16} propose two frameworks (IVIPER and MAVIPER) to extract interpretable coordination policies of MARL in sight of the decision tree. 
Zabounidis et al. \cite{pmlr-v205-zabounidis23a} incorporate interpretable concepts from domain experts into MARL models trained. This approach improves interpretability, allows experts to understand which high-level concepts are used by the policy, and intervenes to improve performance.

MARL for decision transparency involves not only the transparency of single-agent decisions but also the study of complex interactions among multiple agents. Currently, although there have some related research works, it is still relatively small, and more research works are needed to explore how to make MARL more transparent for better application to real-world problems.

\section{Challenges on Human-Compatible Multi-agent Reinforcement Learning} \label{Challenges}
The Human-Cyber-Physical System (HCPS) is developed based on the Cyber-Physical System (CPS) and integrates computer science, automatic technology, communication science, etc \cite{hcps,liuhcps}. The applications of MARL summarized in Section \ref{Applications} of this paper are typical of HCPS.
Humans are seen as an essential component of HCPS. Therefore, the design of MARL algorithms needs to take into account the human factor.
In addition to the challenges of scalability and non-stationary, MARL in HCPS faces many additional challenges due to the interactions between humans, physical systems, and computer systems. 

\subsection{Non-stationarity due to Human Intervention}
Non-stationarity refers to the dynamic changes in the environment or the behavior of agents over time. The existing MARL is based on SG, where the number of agents remains constant during the training process. Currently, research on non-stationarity in MARL is limited to the CPS level, only considering the non-stationarity caused by changes in agent policies on the overall environment\cite{}. However, in HCPS, humans interact continuously with the CPS, and human behavior can affect the dynamic changes in the CPS system. In addition, the reward function for MARL agents is defined by human experts. Human needs will change with social progress, and the reward function for MARL agents will change accordingly. This is also an essential factor causing non-stationarity in HCPS. How to design stable MARL algorithms against human intervention is a vital challenge.
\subsection{Diversity of Human Behavior}
Human behavior is diverse due to the influence of different geographies, cultures, and beliefs. In HCPS, MARL needs to model human behavior in order to better achieve intelligence in interaction with humans. The quality of understanding human behavior predominantly affects the user experience of CPS. For example, in intelligent education, MARL agents need to understand student behavior well to better recommend personalized services for different students. However, the diversity of behavior makes this process very challenging. The current research for modeling human behavior is limited to the societal level and only takes into account human behavior, not the possible influence of machine intelligence on human behavior. How to consider the influence of machines on human behavior in the process of modeling human behavior is a significant challenge.

\subsection{Complex Heterogeneity of HCPS}
The complexity of HCPS manifests itself in various aspects, including human heterogeneity, physical system heterogeneity, cyber system heterogeneity, and temporal heterogeneity. Human heterogeneity refers to the diversity of human behavior and the different roles played by humans in systems with different functions. Physical system heterogeneity refers to the variety of sensors used, such as GPS and cameras in UAV transportation systems. Cyber system heterogeneity is composed of various software, hardware, and algorithms, which require the integration of multiple intelligent algorithms due to the complexity of multi-source information and multi-task decision-making. This cannot be achieved by a single end-to-end algorithm. Finally, temporal heterogeneity is when making decisions; MARL agents require defining different time intervals based on the actual situation at each time step. How to design MARL algorithms to handle the decision-making process of complex heterogeneous HCPS is an enormous challenge. 

\subsection{Scalability of Multi-human and Multi-machine}
HCPS is a complex system of multi-human and multi-machine coexistence. Thus, MARL used for intelligent decision-making should have strong scalability, and the agent here should have a broad concept that includes both humans and intelligent machines. However, as the number of agents increases, the joint action space of agents grows exponentially, which makes the scalability of MARL algorithms poor. Existing research only focuses on the scalability of the number of machines without considering human factors. Designing scalable multi-agent reinforcement learning algorithms that are suitable for complex and heterogeneous HCPS is a significant challenge. 

\section{Conclusion} \label{Conclusion}
This paper summarizes the fundamental methods of MARL and reviews its relevant research in various fields, such as smart transportation, unmanned aerial vehicles, intelligent information system, public health and intelligent medical diagnosis, smart manufacturing, financial trade, network security, smart education, and RL for science. In order to better serve human society, it is necessary to develop a trustworthy MARL. Therefore, we define trustworthy MARL from the perspectives of safety, robustness, generalization, and ethical constraints and summarize the current research and limitations in these areas. 
Finally, we discuss the additional challenges when considering HCPS in MARL, which is crucial for its practical application in human society.
We hope this paper can provide a comprehensive review of various research approaches and application scenarios, encouraging and promoting the application of MARL in human societies for better service to humans.


\bibliographystyle{ACM-Reference-Format}
\bibliography{ref.bib}

\end{document}